\documentclass[10pt,twocolumn,letterpaper]{article}

\usepackage{wacv}              

\definecolor{wacvblue}{rgb}{0.21,0.49,0.74}
\usepackage[pagebackref,breaklinks,colorlinks,allcolors=wacvblue]{hyperref}
\usepackage{multirow}
\usepackage{float}
\usepackage{placeins}
\usepackage{algorithm}
\usepackage{algpseudocode}

\def\wacvPaperID{2142} 
\def\confName{WACV}
\def\confYear{2027}

\title{FSDAM: Few-Shot Driving Attention Modeling via Vision-Language Coupling}

\author{
Kaiser Hamid$^{1}$\quad
Can Cui$^{2}$\quad
Khandakar Ashrafi Akbar$^{3}$\quad
Ziran Wang$^{2}$\quad
Nade Liang$^{1}$\\
{\small $^{1}$Texas Tech University\quad
$^{2}$Purdue University}\\
{\small $^{3}$Towson University}\\
{\tt\small kaiserhamid.munna@ttu.edu}\\\\
{\small \url{https://fsdam-vlc.github.io/}}
}
\begin{document}
\maketitle
\begin{abstract}
Understanding not only where drivers look but also why their attention shifts is 
essential for interpretable human–AI collaboration in autonomous driving. Driver 
attention is not purely perceptual but semantically structured. Thus attention 
shifts can be learned through minimal semantic supervision rather than dense large-
scale annotation. We present \textbf{FSDAM} (\textbf{F}ew-\textbf{S}hot 
\textbf{D}river \textbf{A}ttention \textbf{M}odeling), a framework that achieves 
joint spatial attention prediction and structured explanation generation using 90 
annotated examples. Our key insight is to decompose attention into an explicit 
reasoning representation, including scene context, current focus, anticipated next
focus, and causal explanation, and to learn next-focus anticipation through minimal-
pair supervision. To address task conflict and large sample requirements of 
existing models, and to mitigate task interference under limited data, we introduce
a novel dual-pathway architecture in which separate modules handle spatial 
prediction and caption generation. In addition, we use a training-only vision–
language alignment mechanism that injects semantic priors into spatial learning 
without increasing inference complexity, mitigating task interference under few-
shot training. Despite extreme data scarcity, FSDAM achieves competitive 
performance in gaze prediction, and generates coherent, context-aware structural 
reasoning for improved interpretability. The model further demonstrates strong zero-
shot generalization across multiple driving benchmarks. These results suggest that 
semantically grounded attention modeling enables data-efficient learning and 
provides a scalable path toward explainable driver attention systems in data-
constrained environments.
\end{abstract}
    
\enlargethispage{2\baselineskip}
\section{Introduction}
\label{sec:intro}
\begin{figure*}[t]
    \centering
    \includegraphics[width=0.92\textwidth]{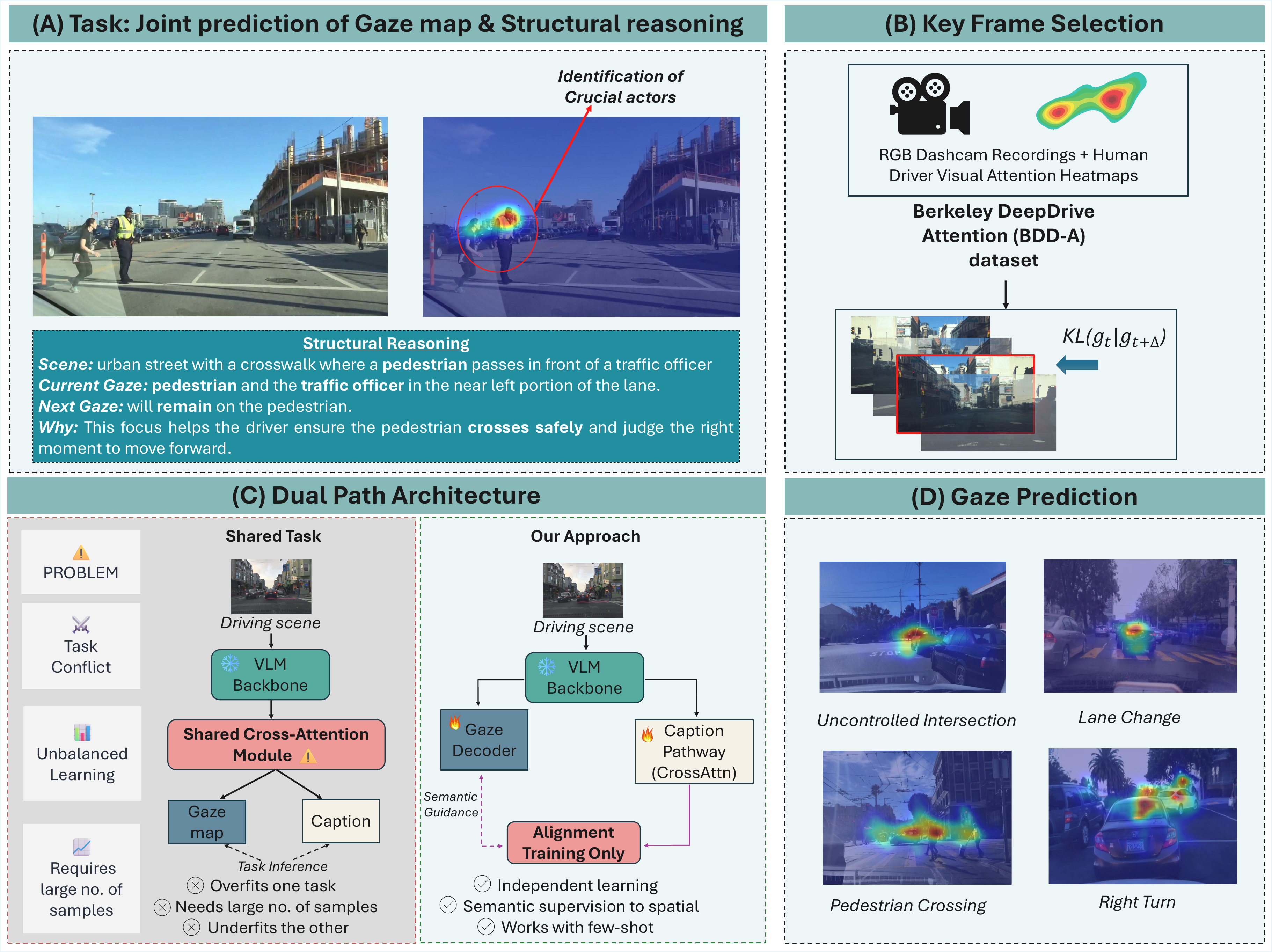}
    \caption{\textbf{Few-Shot Driving Attention Modeling (FSDAM).}
    \textbf{(A)}~Joint prediction of a gaze map and structured reasoning
    (\textit{Scene, Current Gaze, Next Gaze, Why}).
    \textbf{(B)}~Key-frame selection on BDD-A via KL-divergence mining to select
    high-change gaze-transition moments between frame pairs.
    \textbf{(C)}~Dual-pathway design with separate gaze and caption pathways,
    coupled through a \emph{training-only} vision--language alignment loss.
    \textbf{(D)}~Qualitative gaze predictions across four representative driving scenarios.}
    \label{fig:intro}
\end{figure*}
Driving is a fundamentally visual and anticipatory task. Drivers continuously allocate attention across traffic signals, pedestrians, vehicles, and road geometry, not only passively reacting to static observation but actively anticipating potential hazards and right-of-way conflicts. This selective attention critically shapes situation awareness (SA) and downstream decision-making. Suboptimal attention allocation with critical actors overlooked remains a major contributor to traffic accidents, with high crash rates at intersections and during lane changes  \cite{9312486,10588743}. Understanding where and why drivers allocate attention is therefore essential not only for safety analysis but also for autonomous vehicles (AVs) operating in mixed-traffic environments. Furthermore, modeling driver attention enables cross-verification between human intent and machine decisions in human-AI collaboration, supporting interpretable and trustworthy autonomy. 

Despite its importance, modeling driver attention remains challenging. Most existing approaches formulate attention prediction as a supervised saliency estimation problem, learning spatial gaze distributions from large-scale eye-tracking datasets. These methods depend heavily on densely annotated data collected under well-represented conditions to stabilize low-level perception \cite{li2025recogdrive}. However, large-scale gaze data collection is costly, privacy-sensitive, and often lacks diversity in safety-critical or rare scenarios \cite{kroger2020does,ghosh2025roadwork}. Moreover, the combinatorial growth of traffic configurations makes exhaustive dataset expansion increasingly impractical \cite{ghosh2025roadwork,li2025recogdrive}. This reliance on extensive supervision limits scalability and generalization to new domains.

Beyond data scarcity, a more fundamental limitation lies in how attention is conceptualized. Treating driver attention as static perceptual saliency overlooks its inherently forward-looking and semantically structured nature. Human drivers do not simply fixate on visually prominent regions; they anticipate future events based on traffic rules, agent behaviors, and scene context. Recent vision–language models (VLMs) offer a promising direction by injecting semantic reasoning into attention modeling. Works such as DriveLM \cite{sima2023drivelm}, LLada \cite{zhou2025where}, and GazeXplain \cite{chen2024gazexplainlearningpredictnatural} demonstrate that coupling spatial attention with language reasoning improves interpretability and performance. However, these approaches typically require tens of thousands of annotated samples to jointly align gaze and language representations, limiting their applicability in data-constrained scenarios.

To address these limitations, we develop \textbf{FSDAM} (Figure \ref{fig:intro}), a framework that jointly learns gaze prediction and attention-grounded caption generation from 90 annotated examples. To our knowledge, \textit{this is the first work showing that attention-conditioned language generation can be learned effectively in a few-shot regime}. We also introduce a next-attention target prior derived from semantic cues via minimal-pair supervision. Unlike motion-based temporal forecasting, our method estimates future gaze positions from a single frame, conditioned on the current scene. 
We construct supervision by selecting frame pairs with clear attention shifts and annotating each transition with a structured caption describing scene context, current focus, anticipated next focus, and causal rationale. At inference, the model  captures the forward-looking nature of human attention without requiring video input. Our key insight is to couple spatial attention and language understanding as complementary supervision signals through an explicit, structured attention-reasoning format. Building on this design, we propose a \textbf{dual-pathway architecture} that decouples gaze prediction from caption generation, enabling effective joint learning under extreme data scarcity. In summary, we make the following contributions:

\begin{itemize}
    \item \textbf{First few-shot approach for attention-based generation.} We are \textbf{the first of its kind} to achieve joint spatial attention prediction and natural language explanation in a few-shot learning regime, training from 90 examples, two orders of magnitude more data efficient than existing joint modeling approaches \cite{zhou2025where,chen2024gazexplainlearningpredictnatural}.
    \item \textbf{Structured attention reasoning formulation.} We reformulate driver attention modeling as structured anticipation rather than static saliency prediction. Specifically, we decompose attention into four explicit components, including scene context, current focus, anticipated next focus, and causal explanation, establishing a semantic representation that links spatial gaze patterns with forward-looking reasoning.
    \item \textbf{Dual-pathway architecture with training-only alignment.} We propose a decoupled gaze–language architecture that mitigates negative transfer under limited data. A training-only vision–language alignment mechanism injects semantic priors into spatial prediction without increasing inference complexity.
    \item \textbf{Data-efficient and transferable attention modeling.} Despite training in a few-shot regime, our method achieves competitive performance against fully supervised baselines trained on more data and demonstrates strong zero-shot generalization across multiple driver attention benchmarks.
\end{itemize}
\section{Related Works}
\subsection{Driver Attention Modeling}
Driver attention modeling has been studied as a visual saliency prediction task, estimating spatial gaze heatmaps from dashcam images 
or video sequences~\cite{zhou2025where}. Early works applied CNNs to learn correlations between scene features and eye fixations, 
while subsequent models incorporated temporal dynamics and multi-modal inputs using CNN-LSTMs and feature fusion from RGB, optical 
flow, and semantic segmentation~\cite{7789504,10.1007/978-3-030-20873-8_42}. The SEEV model~\cite{steelman2017theory} outlines bottom-
up (salience, effort) and top-down (expectancy, value) factors influencing attention allocation. Existing models typically emphasize 
either low-level visual features or task-level signals such as GPS and traffic semantics~\cite{8751968,10588743}. These approaches 
provide limited insights into the underlying causes of attention allocation beyond saliency.

\subsection{Vision-Language Models for Explainable Attention}
Recent work has applied VLMs to enhance attention interpretability through natural language. “Attention Neural Baby 
Talk"~\cite{8917187} generates captions highlighting hazardous elements by aligning attention masks with descriptions, while 
DRAMA~\cite{malla2023drama} pairs videos with QA annotations explaining risk rationale. Large-scale VLMs like BLIP~\cite{li2022blip},
Flamingo~\cite{alayrac2022flamingo}, and LLaVA~\cite{10.5555/3666122.3667638} have enabled new approaches through vision-language 
pretraining and few-shot learning 
capabilities~\cite{xu2024drivegpt4interpretableendtoendautonomous,DriveVLM,zhou2024visionlanguagemodelsautonomous}.

For driver attention specifically, LLada~\cite{zhou2025where} jointly models attention maps and text through a shared cross-attention module trained on W3DA ($\sim$70k samples), and GazeXplain~\cite{chen2024gazexplainlearningpredictnatural} aligns scanpaths with explanations for general visual attention. Both depend on large-scale joint supervision. In contrast, FSDAM decouples the gaze and caption pathways and uses a training-only alignment loss that adds no inference cost, enabling few-shot learning from 90 samples.

\subsection{Few-Shot Learning and Data-Efficient Adaptation}
Few-shot learning addresses generalization from minimal examples. In semantic segmentation, PANet~\cite{wang2020panetfewshotimagesemantic} achieves 48.1\% mIoU with 5 images per class, while few-shot object detection methods report 15-30\% AP with 10-30 examples~\cite{kang2019fewshotobjectdetectionfeature,fan2020fewshotobjectdetectionattentionrpn,yan2024anomalysdfewshotmulticlassanomaly}. In generative modeling, DreamBooth~\cite{ruiz2023dreamboothfinetuningtexttoimage} personalizes Stable Diffusion using 3-5 images (CLIP-I 0.74), while UFC~\cite{kim2023universalfewshotlearningdense} achieves 87.3\% accuracy with 30 examples versus 89.1\% for fully-supervised baselines (10K+ examples).

Parameter-efficient fine-tuning enables adaptation with minimal updates. LoRA~\cite{hu2022lora} matches full fine-tuning performance 
while updating 0.01\% of parameters, reducing GPT-3's trainable parameters from 175B to 4.7M. For VLMs, LoRA adaptation maintains 
95\%+ performance with 10-20M trainable parameters~\cite{houlsby2019parameterefficienttransferlearningnlp}. Flamingo demonstrates in-
context learning, improving VQAv2 from 49.2\% (0-shot) to 63.1\% (32-shot)~\cite{alayrac2022flamingo}, though this approach shows 
high variance and struggles with structured outputs like spatial maps~\cite{dong2024surveyincontextlearning}.

In autonomous driving, few-shot learning remains underexplored. Most attention systems train on hundreds of thousands of frames~\cite{palazzi2018predicting,xia2018predictingdriverattentioncritical,fang2019dada2000drivingaccidentpredicted}. 

\begin{figure*}[t]
  \centering
  \includegraphics[width=0.99\linewidth]{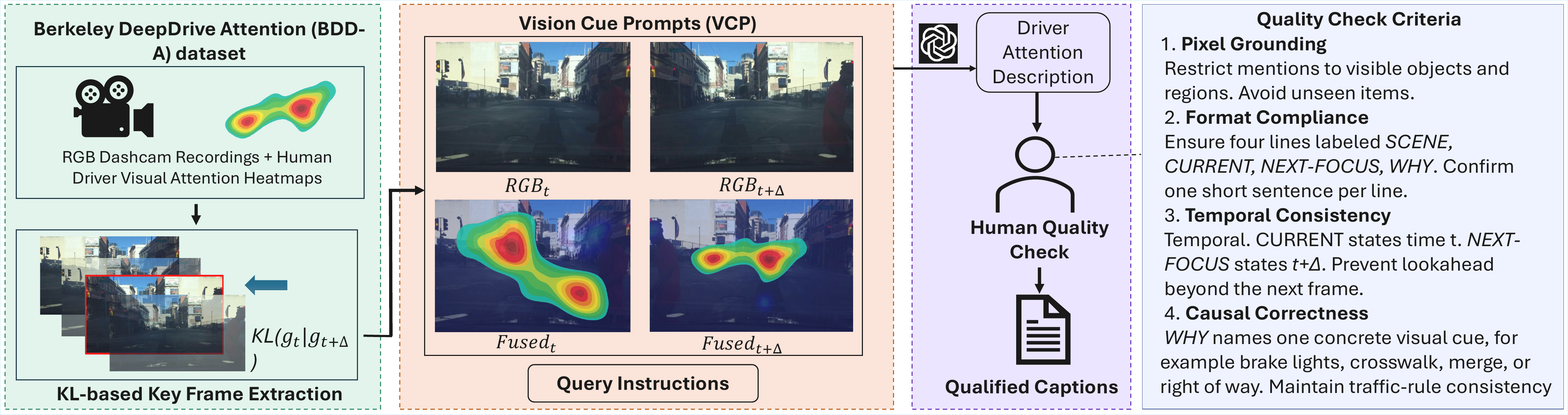}
  \caption{Dataset curation pipeline. From BDD-A~\cite{10.1007/978-3-030-20873-8_42} videos to paired frames, GPT-4o captioning with fixed template, human verification, and final captions. Full details in supplementary material.}
  \label{fig:dataset_curation}
\end{figure*}

\section{Method}
\begin{figure}[t]
  \centering
  \includegraphics[width=1\linewidth]{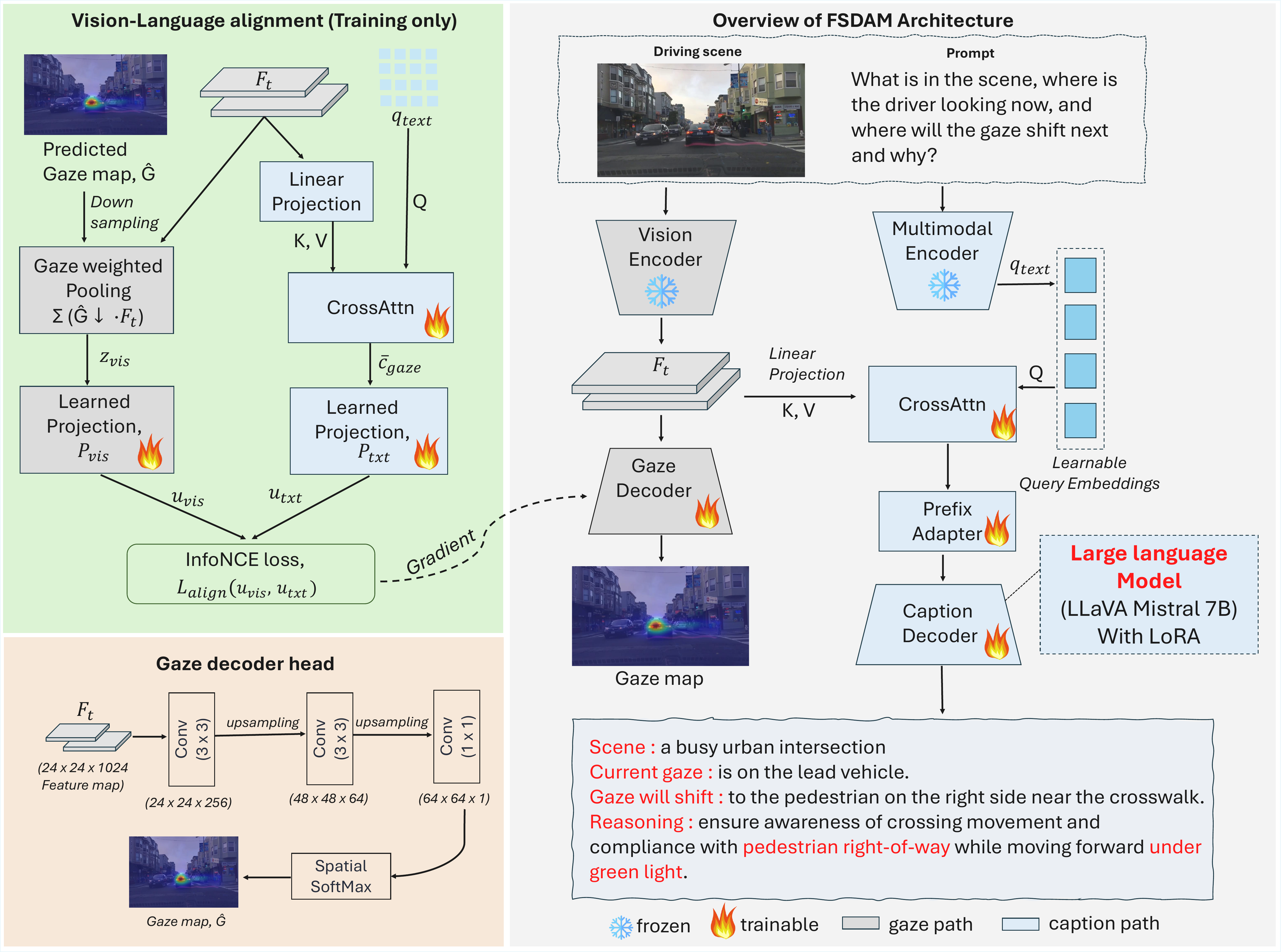} 
  \caption{Overview of the proposed FSDAM architecture (right). A frozen vision--language backbone extracts spatial features $F_t$ and text embeddings $q_{\text{text}}$. The gaze pathway predicts the attention map $\widehat{G}$, while the explanation pathway performs cross-attention over $F_t$ to generate structured reasoning. The training-only vision--language alignment module (top left) applies contrastive supervision between gaze-weighted pooled visual features and text features, and backpropagates gradients to the gaze decoder. The gaze decoder head (bottom left) upsamples $F_t$ to produce $\widehat{G}$.}
  \label{fig:architecture}
\end{figure}

We present our few-shot dependent driver attention modeling framework that jointly models spatial driver attention prediction and 
structured natural language explanation. These two tasks depend on totally different kinds of spatial understanding, where captioning 
needs global semantic reasoning while gaze prediction requires localized spatial sensitivity. A shared cross-attention module tends 
to overfit one task and underfit another~\cite{Li_2024,fifty2021efficientlyidentifyingtaskgroupings}. To address this imbalance, we 
introduce a dual-pathway architecture (Figure \ref{fig:architecture}) in which gaze prediction and caption generation are handled by 
separate modules while still leveraging shared visual features. A vision-language alignment mechanism further provides semantic 
supervision to spatial prediction, ensuring predicted attention regions correspond to meaningful visual content. This design enables 
effective joint learning despite data scarcity.

\subsection{Problem Formulation}

Given a driving scene image $I \in \mathbb{R}^{H \times W \times 3}$, we jointly predict: (1)~a spatial attention distribution $\widehat{G} \in \Delta^{S \times S}$ indicating where the driver looks, where $\Delta^{S \times S}$ denotes the probability simplex over an $S{\times}S$ grid with $S{=}64$, and (2)~a structured attention reasoning representation $\widehat{y}$ following the format:
\begin{equation}
    \widehat{y} = (C_{scene},C_{current}, C_{next}, C_{why})
\end{equation}

Here, each $C_*$ is a natural language sentence describing one aspect of the attention state. Specifically, $C_{{scene}}$ summarizes 
the global scene context, $C_{current}$ describes the driver's present focus of attention, $C_{\text{next}}$ specifies the likely 
next focus of attention (``will check the crosswalk''), and $C_{why}$ provides the causal rationale underlying this transition. 
Together, the four components form a structured attention reasoning representation that links spatial attention patterns with high-
level semantic explanations, enabling the model to learn gaze behavior from single frames.


Our architecture builds upon LLaVA-Next-1.6~\cite{liu2024llavanext}, which comprises a CLIP-ViT-L/14 vision 
encoder~\cite{radford2021learning} extracting spatial features $F_t \in \mathbb{R}^{B \times 1024 \times 24 \times 24}$ 
for input frame $t$, and a Mistral-7B language model~\cite{jiang2023mistral7b} producing image-conditioned text embeddings 
$q_{\text{text}} \in \mathbb{R}^{B \times d_{\ell}}$, where $d_{\ell}$ is the language model's hidden dimension, through multimodal 
fusion~\cite{alayrac2022flamingo,liu2024llavanext}. These frozen features feed into specialized pathways for gaze prediction and 
caption generation, with task-specific adaptation enabled by LoRA~\cite{hu2022lora}. We optimize 13.6M task-specific parameters 
(0.2\% of the 7B backbone) while keeping pretrained components frozen to prevent overfitting.

\subsection{Spatial Attention Prediction}
The gaze pathway predicts where drivers look. A convolutional decoder upsamples $F_t$ from $24{\times}24$ to $64{\times}64$ 
resolution through successive convolution and bilinear upsampling 
layers~\cite{long2015fullyconvolutionalnetworkssemantic,ronneberger2015unetconvolutionalnetworksbiomedical}, then applies spatial 
softmax to produce $\widehat{G} \in \Delta^{S \times S}$. We supervise with forward KL divergence to encourage covering all human
fixation regions:
\begin{equation}
\mathcal{L}_{\text{KL}} = \text{KL}(G \| \widehat{G}) = \sum_{i,j} G(i,j) \log \frac{G(i,j)}{\widehat{G}(i,j)}.
\end{equation}
Here, $G$ is the ground-truth gaze distribution and $i, j$ index spatial positions over the $S{\times}S$ grid. This alone produces overly diffuse predictions under limited data. We augment with a \emph{blur-gap regularizer} that enforces spatial 
sharpness by penalizing predictions whose quality does not degrade 
under Gaussian smoothing ($\sigma{=}1.0$). Specifically, we smooth 
$\widehat{G}$ to obtain $\widetilde{G}$, then penalize cases where 
blurring does not degrade the prediction:
\begin{equation}
\mathcal{L}_{\text{gaze}} = \mathcal{L}_{\text{KL}} + \lambda \cdot \max\big(0, \text{KL}(G \| \widetilde{G}) - \mathcal{L}_{\text{KL}} + \epsilon\big),
\end{equation}
with $\lambda{=}0.3$ and $\epsilon{=}0.05$. This encourages confident spatial localization by exploiting the sharpness--blur 
relationship in saliency maps~\cite{kümmerer2016deepgazeiireadingfixations}, where diffuse predictions degrade minimally under 
Gaussian smoothing while sharp, well-localized predictions do not. The gaze pathway receives gradients only from 
$\mathcal{L}_{\text{gaze}}$ during forward passes, with semantic supervision added through alignment (Section~\ref{sec:align}).

\subsection{Attention-Grounded Caption Generation}

Following prefix-tuning~\cite{li2021prefixtuning}, we expand $q_{\text{text}}$ into $M$ queries via projection $W_{\text{cap}}$, then perform cross-attention~\cite{vaswani2017attention} over $F_t$:
\begin{equation}
Q = W_{\text{cap}}(q_{\text{text}}) \in \mathbb{R}^{B \times M \times d}, \quad \text{CTX} = \text{CrossAttn}(Q, K, V).
\end{equation}
Mean pooling aggregates $M$ context vectors into $\bar{c}_{\text{cap}}$, which a Prefix Adapter $\Psi$ projects to visual prefix tokens $P = \Psi(\bar{c}_{\text{cap}})$ conditioning the decoder. We supervise with autoregressive cross-entropy:
\begin{equation}
\mathcal{L}_{\text{cap}} = -\sum_{t=1}^{T} \log p_\theta(y_t \mid y_{<t}, P, I),
\end{equation}
where $y$ contains scene description, current attention, next focus of attention, and causal reasoning. LoRA~\cite{hu2022lora} enables task-specific tuning while keeping most parameters frozen.

\subsection{Vision-Language Alignment}
\label{sec:align}

To ensure predicted attention regions align with semantic content, we introduce a training-only alignment mechanism. A dedicated cross-attention block processes text queries $q_{\text{text}}$ to produce text-conditioned features $\bar{c}_{\text{gaze}}$. In parallel, we pool spatial features $F_t$ using predicted gaze $\widehat{G}_{\downarrow}$ (downsampled to match $F_t$ resolution) as soft weights:
\begin{equation}
z_{\text{vis}} = \sum_{i,j} \widehat{G}_{\downarrow}(i,j)\, F_t[:, :, i, j].
\end{equation}
Here, using the model's own predicted attention as pooling weights maintains consistency between training and inference~\cite{chen2020simpleframeworkcontrastivelearning,grill2020bootstrap}.

Both representations project into a shared 256-dimensional space via learned projections $P_{\text{vis}}$ and $P_{\text{txt}}$, producing $u^{\text{vis}} = P_{\text{vis}}(z_{\text{vis}})$ and $u^{\text{txt}} = P_{\text{txt}}(\bar{c}_{\text{gaze}})$. We align these with InfoNCE contrastive loss~\cite{oord2018representation}:
\begin{equation}
\mathcal{L}_{\text{align}}
= -\frac{1}{B} \sum_{i=1}^{B}
\log
\frac{\exp\big(\text{sim}(u_i^{\text{vis}}, u_i^{\text{txt}})/\tau\big)}
{\sum_{j=1}^{B} \exp\big(\text{sim}(u_i^{\text{vis}}, u_j^{\text{txt}})/\tau\big)},
\end{equation}
where $\text{sim}(\cdot,\cdot)$ denotes cosine similarity and $\tau$ is a temperature parameter. Gradients pass through spatial pooling to the gaze decoder, pushing attention toward semantically relevant regions.

\subsection{Training and Inference}

We jointly optimize all three objectives through a weighted combination:
\begin{equation}
\mathcal{L}=w_g\,\mathcal{L}_{\text{gaze}}+w_c\,\mathcal{L}_{\text{cap}}+w_a\,\mathcal{L}_{\text{align}},
\label{eq:total_loss}
\end{equation}
where $(w_g,w_c,w_a)=(1.0,1.0,0.2)$ balance gaze prediction, caption generation, and vision-language alignment. The weights were chosen through a small-scale grid search on a validation split and were stable across experiments.

\textbf{Training.}
We train using AdamW optimizer with separate learning rates for LoRA and task modules. 
Training uses mixed precision, gradient accumulation for an effective batch of 16, and input size $336{\times}336$. 
Only LoRA and lightweight adapters are updated, totaling 13.6M trainable parameters ($\sim$0.2\% of the backbone). 
Learning rate is adjusted via a ReduceLROnPlateau scheduler.

\textbf{Inference.}
During inference, only gaze and caption branches are active. 
The gaze decoder predicts $\widehat{G}$ from visual features $F_t$, while the caption pathway generates structured explanations through cross-attention and prefix adaptation. 
The alignment module is used only during training and adds no cost at inference.

\section{Experimental Setup}

\textbf{Dataset Preparation.}
We construct a gaze-language dataset from BDD-A dataset~\cite{10.1007/978-3-030-20873-8_42}, which contains braking event videos 
selected from large-scale, crowd-sourced driving video data combined with compiled human gaze data from 6.5 seconds prior and 3.5 
seconds after each braking event. To identify frames capturing meaningful attention transitions, we employ a KL-divergence-based 
selection algorithm~\cite{shlens2014noteskullbackleiblerdivergencelikelihood} that detects moments of maximum gaze distribution 
change between consecutive frames. Local peaks in the KL divergence curve mark anchor frames $t$ where attention shifts abruptly. 
For each anchor, we select a target frame $t+\Delta$ within $[3,18]$ frames that maximizes divergence from the anchor, capturing the 
strongest future attention transition. Following temporal sampling strategies from video action 
recognition~\cite{carreira2018quovadisactionrecognition}, we filter clips shorter than 50 frames and retain the top-K = 2 anchor–
target pairs per video for temporal diversity. Each resulting sample $(I_t, I_{t+\Delta}, G_t, G_{t+\Delta})$ consists of two frames 
and their corresponding gaze maps. We generate structured captions via GPT-4o~\cite{openai2024gpt4technicalreport} with human 
verification (Figure~\ref{fig:dataset_curation}), yielding 90 training examples systematically sampled from the candidate key frame 
pairs to evenly cover eight driving scenario categories (see supplementary for the taxonomy and category-wise counts).

\textbf{Implementation Details.}
All experiments are implemented in PyTorch~\cite{paszke2019pytorchimperativestylehighperformance}. We build on the LLaVA-Next Mistral-
7B backbone, which integrates a CLIP-based vision tower pretrained on large-scale image–text pairs. The vision–language backbone 
remains frozen during training, and only lightweight task-specific components 
(gaze decoder, prefix adapter, alignment head, and LoRA adapters) are updated. Training is performed on a single NVIDIA GH200 GPU 
with input resolution $336{\times}336$, 
batch size 4, and gradient accumulation over 4 steps (effective batch 16). We use the \textit{AdamW} optimizer with learning rates of 
$1{\times}10^{-4}$ for LoRA and $2{\times}10^{-4}$ for task heads, a \textit{ReduceLROnPlateau} scheduler (factor 0.5, patience 2), 
and mixed precision of \textit{bfloat16}.

\textbf{Evaluation Metrics.}
We evaluate our model using standard saliency and captioning metrics. For gaze prediction, we adopt CC, KL, SIM, AUC-J, AUC-B, and 
NSS following the official MIT Saliency Benchmark~\cite{bylinskii2017differentevaluationmetricstell}. For caption generation, we 
report BLEU, METEOR, ROUGE-L, CIDEr-R, and BERTScore using the COCO Caption Evaluation 
Toolkit~\cite{chen2015microsoftcococaptionsdata}. All metrics are computed using their standardized implementations to ensure fair 
comparison with prior work.
\section{Results \& Analysis}
\subsection{Baselines}
We evaluate gaze prediction on four datasets: BDD-A~\cite{10.1007/978-3-030-20873-8_42}, DADA-2000~\cite{fang2023dadadriverattentionprediction}, DR(eye)VE~\cite{palazzi2018predicting}, and W3D~\cite{zhou2025where}. We compare against classical saliency/gaze baselines (U\textsuperscript{2}-Net~\cite{Qin_2020_PR}, MINet~\cite{pang2020multiscaleinteractivenetworksalient}, DBNet~\cite{9812524}, DeepLabV3~\cite{chen2017rethinkingatrousconvolutionsemantic}) and a joint gaze--language baseline (LLada~\cite{zhou2025where}). Unless noted otherwise, reproduced baselines are trained on the full BDD-A training split using the official implementations and evaluated on each dataset's official test split. To provide a fair few-shot reference across both classical saliency and joint gaze–language baselines, we retrain U\textsuperscript{2}-Net and LLada on the same 90-sample BDD-A subset as FSDAM (Table~\ref{tab:bdda_indomain}, \emph{Few-shot}), ensuring differences are not attributable solely to training-data scale.

\begin{table}[t]
\centering
\caption{In-domain driver attention prediction on BDD-A. \textbf{Bold} = best,
\underline{underline} = second best. $^*$ results from original paper (trained on BDD-A).
$^\dagger$ fine-tuned on 90 BDD-A samples.}
\resizebox{0.9\columnwidth}{!}{
\begin{tabular}{llcccccc}
\toprule
\textbf{Method} & \textbf{Training Data} &
CC$\uparrow$ & KL$\downarrow$ & SIM$\uparrow$ & AUC-J$\uparrow$ & AUC-B$\uparrow$ & NSS$\uparrow$ \\
\midrule
\multicolumn{8}{l}{\textit{Fully Supervised (Full BDD-A training set)}} \\
\midrule
U²-Net~\cite{Qin_2020_PR}
  & Full BDD-A & 0.56 & 1.50 & \textbf{0.47} & 0.94 & \underline{0.88} & 3.95 \\
MINet~\cite{pang2020multiscaleinteractivenetworksalient}
  & Full BDD-A & 0.46 & 10.30 & 0.16 & 0.89 & 0.82 & 3.67 \\
DBNet~\cite{9812524}
  & Full BDD-A & \underline{0.57} & 1.30 & 0.41 & 0.95 & \textbf{0.91} & \textbf{4.41} \\
DeepLabV3~\cite{chen2017rethinkingatrousconvolutionsemantic}
  & Full BDD-A & 0.47 & 9.62 & 0.21 & \textbf{0.97} & 0.75 & 2.56 \\
LLada$^*$~\cite{zhou2025where}
  & Full BDD-A & \textbf{0.60} & \underline{1.16} & \textbf{0.47} & -- & -- & -- \\
\midrule
\multicolumn{8}{l}{\textit{Few-Shot (90 BDD-A samples)}} \\
\midrule
U²-Net$^\dagger$~\cite{Qin_2020_PR}
  & BDD-A-90 & 0.10 & 3.43 & 0.13 & 0.74 & 0.68 & 0.63 \\
LLada$^\dagger$~\cite{zhou2025where}
  & BDD-A-90 & 0.37 & 1.86 & 0.32 & 0.91 & 0.81 & 2.91 \\
\textbf{FSDAM (Ours)}
  & BDD-A-90 & \textbf{0.60} & \textbf{1.13} & \underline{0.43} & \underline{0.96} & \textbf{0.91} & \underline{4.10} \\
\bottomrule
\end{tabular}}
\label{tab:bdda_indomain}
\end{table}

For captioning, we compare on W3D against fully supervised baselines (GazeXplain~\cite{chen2024gazexplainlearningpredictnatural}, LLada~\cite{zhou2025where}; ${\sim}$70k samples), zero-shot and in-context models (Qwen-VL~\cite{bai2025qwen25vltechnicalreport}, LLaVA~\cite{10.5555/3666122.3667638}), and few-shot two-stage baselines pairing gaze models with LLaVA (DeepGazeIIE~\cite{linardos2021deepgazeiiecalibratedprediction}+LLaVA, MLNet~\cite{8356626}+LLaVA), trained on our same 90 BDD-A samples. Additional analyses and experiments are provided in the supplementary material.

\subsection{Quantitative Analysis}
\subsubsection{Spatial Attention Prediction}
Table~\ref{tab:bdda_indomain} and Table~\ref{tab:zero_shot_transfer} report quantitative results for driver attention prediction across four datasets. Table~\ref{tab:bdda_indomain} evaluates in-domain performance on BDD-A under both fully supervised and few-shot regimes. Table~\ref{tab:zero_shot_transfer} evaluates zero-shot transfer to DADA-2000, DR(eye)VE, and W3D, where all baselines are trained on full BDD-A and FSDAM uses only 90 samples. All reproduced models were trained on BDD-A for consistency. FSDAM provides a strong balance between sample efficiency and predictive accuracy. With only 90 training samples, it reaches performance that aligns with or exceeds models trained on much larger datasets, including W3D with $\sim$70K frames, DADA-2000 with 658K frames, and DR(eye)VE with 555K frames.

\textbf{In-domain Table~\ref{tab:bdda_indomain}.} On BDD-A\cite{10.1007/978-3-030-20873-8_42}, FSDAM obtains the lowest KL divergence at 1.13. This improves over LLada\cite{zhou2025where} by 2.6\% and over DBNet\cite{9812524} by 13\%. It matches LLada\cite{zhou2025where} in CC at 0.60 and ranks second in NSS at 4.10, indicating better spatial attention prediction. Compared to the fully supervised MINet\cite{pang2020multiscaleinteractivenetworksalient}, FSDAM improves CC by 30\% and reduces KL divergence by 89\%, which highlights its effectiveness with limited data. In the few-shot setting, classical saliency models such as U²-Net$^\dagger$~\cite{Qin_2020_PR} degrade severely to CC=0.10, confirming that appearance-based features cannot generalize under extreme data scarcity. LLada$^\dagger$~\cite{zhou2025where}, despite sharing the same VLM backbone, also drops substantially to CC=0.37 and KL=1.86, demonstrating that vision-language alignment alone is insufficient without our dual-pathway few-shot design.

\textbf{Cross-dataset transfer Table~\ref{tab:zero_shot_transfer}.} Our cross-dataset evaluation demonstrates the strong transferability of our approach. On DR(eye)VE\cite{palazzi2018predicting}, FSDAM reaches the lowest KL divergence at 0.77, improving over LLada\cite{zhou2025where} by 26\% and DBNet\cite{9812524} by 57\%. It also achieves the highest SIM at 0.54, a 20\% gain over U²-Net\cite{Qin_2020_PR}. This shows that vision-language alignment helps capture domain stable attention cues that transfer to highway scenarios.

On DADA-2000\cite{fang2023dadadriverattentionprediction}, FSDAM maintains competitive performance while using far fewer samples, achieving the best KL divergence at 1.64 with a 10\% gain over LLada\cite{zhou2025where} and a 13\% gain over DBNet\cite{9812524}. On W3D, FSDAM matches DeepLabV3\cite{chen2017rethinkingatrousconvolutionsemantic}  in CC at 0.53 and beats LLada in KL divergence at 1.27 by 13\%. It also achieves the second best SIM at 0.44 and NSS at 3.50.

Together, these results reveal that strong driver attention modeling does not require exhaustive gaze supervision. FSDAM consistently matches or outperforms fully supervised baselines across diverse driving scenarios, indicating that our dual-pathway vision-language design captures transferrable, semantically grounded attention cues absent from data-intensive saliency models.


\begin{table*}[t]
\centering
\caption{Zero-shot cross-dataset transfer for driver attention prediction. All baselines
are trained on the full BDD-A training set; FSDAM uses only 90 BDD-A samples.
No target-domain fine-tuning is applied for any method.
\textbf{Bold} = best, \underline{underline} = second best.}
\resizebox{\textwidth}{!}{
\begin{tabular}{llcccccccccccccccccc}
\toprule
\multirow{2}{*}{\textbf{Method}} &
\multirow{2}{*}{\textbf{Training Data}} &
\multicolumn{6}{c}{\textbf{DADA-2000}} &
\multicolumn{6}{c}{\textbf{DR(eye)VE}} &
\multicolumn{6}{c}{\textbf{W3D}} \\
\cmidrule(lr){3-8} \cmidrule(lr){9-14} \cmidrule(lr){15-20}
& & CC$\uparrow$ & KL$\downarrow$ & SIM$\uparrow$ & AUC-J$\uparrow$ & AUC-B$\uparrow$ & NSS$\uparrow$
  & CC$\uparrow$ & KL$\downarrow$ & SIM$\uparrow$ & AUC-J$\uparrow$ & AUC-B$\uparrow$ & NSS$\uparrow$
  & CC$\uparrow$ & KL$\downarrow$ & SIM$\uparrow$ & AUC-J$\uparrow$ & AUC-B$\uparrow$ & NSS$\uparrow$ \\
\midrule
U²-Net~\cite{Qin_2020_PR}
  & Full BDD-A
  & 0.42 & 2.18 & \textbf{0.37} & 0.91 & 0.82 & 2.73
  & 0.57 & 1.52 & 0.45 & \underline{0.90} & 0.82 & 3.20
  & 0.44 & 2.10 & \underline{0.37} & 0.90 & 0.81 & 2.81 \\
MINet~\cite{pang2020multiscaleinteractivenetworksalient}
  & Full BDD-A
  & 0.32 & 10.45 & 0.14 & 0.82 & 0.73 & 2.02
  & 0.44 & 8.87  & 0.36 & 0.84 & 0.80 & 2.65
  & 0.35 & 10.51 & 0.14 & 0.83 & 0.77 & 2.54 \\
DBNet~\cite{9812524}
  & Full BDD-A
  & 0.41 & 1.89 & 0.29 & \underline{0.93} & \underline{0.85} & \textbf{3.08}
  & 0.48 & 1.79 & 0.29 & \textbf{0.91}    & \textbf{0.85}    & \underline{3.71}
  & 0.47 & 1.77 & 0.33 & \underline{0.93} & 0.86             & 3.50 \\
DeepLabV3~\cite{chen2017rethinkingatrousconvolutionsemantic}
  & Full BDD-A
  & \underline{0.42} & 10.26 & 0.19 & \textbf{0.96} & 0.73 & 2.23
  & \textbf{0.67} & 8.78 & \underline{0.47} & \textbf{0.91} & 0.84 & 2.74
  & \textbf{0.53} & 9.70 & 0.32 & \textbf{0.95} & 0.81 & 2.35 \\
\midrule
\textbf{FSDAM (Ours)}
  & BDD-A-90 (few-shot)
  & \textbf{0.44} & \textbf{1.64} & 0.33 & 0.92 & \textbf{0.86} & \underline{2.95}
  & \underline{0.61} & \textbf{0.77} & \textbf{0.54} & \textbf{0.91} & \underline{0.84} & \textbf{4.04}
  & \underline{0.53} & \textbf{1.27} & \textbf{0.44} & \underline{0.91} & \underline{0.87} & \underline{3.50} \\
\bottomrule
\end{tabular}}
\label{tab:zero_shot_transfer}
\end{table*}
\subsubsection{Caption Prediction Analysis}
Table~\ref{tab:w3d_caption_comparison} presents caption generation performance across three driving categories on the W3D dataset. Despite training under our few-shot regime, FSDAM achieves results approaching fully trained baselines trained on the full W3D dataset. In normal driving scenarios, FSDAM attains BLEU 0.42, METEOR 0.37, and CIDEr-R 0.83, closely approaching LLada (BLEU 0.44, METEOR 0.36, CIDEr-R 0.96) trained on $\sim$70k samples. Generalization holds across challenging scenarios, with competitive scores in safety-critical situations (BLEU: 0.35, METEOR: 0.33, CIDEr-R: 0.47) and traffic accidents (BLEU: 0.33, METEOR: 0.35, CIDEr-R: 0.84). Among few-shot baselines, two-stage approaches (DeepGazeIIE~\cite{linardos2021deepgazeiiecalibratedprediction}+LLaVA, MLNet~\cite{8356626}+LLaVA) achieve BLEU scores no higher than 0.26 across all scenarios, while FSDAM consistently exceeds 0.33 across normal driving, safety-critical, and accident scenarios. Zero-shot and in-context models remain limited across all scenarios (BLEU~<~0.21), confirming that task-specific grounding is necessary for reliable driver caption generation.

\begin{table*}[t]
\centering
\caption{Comparison of captioning performance across driving scenarios on W3D dataset. 
Fully-trained baselines use the original W3D training data, 
Zero-shot and ICL models require no fine-tuning, 
and our FSDAM and few-shot baselines are trained on a 90-sample BDD-A subset. 
Higher is better.}
\resizebox{\textwidth}{!}{
\begin{tabular}{l|l|cccc|cccc|cccc}
\toprule
\multirow{2}{*}{\textbf{Method}} &
\multirow{2}{*}{\textbf{Training Regime}} &
\multicolumn{4}{c|}{\textbf{Normal Driving}} &
\multicolumn{4}{c|}{\textbf{Safety-Critical Situation}} &
\multicolumn{4}{c}{\textbf{Traffic Accident}} \\
\cmidrule(lr){3-14}
 &  & BLEU & METEOR & ROUGE & CIDEr-R 
    & BLEU & METEOR & ROUGE & CIDEr-R 
    & BLEU & METEOR & ROUGE & CIDEr-R \\
\midrule
\multicolumn{14}{l}{\textit{Fully Trained on W3D}} \\
\midrule
GazeXplain\textsuperscript{*}~\cite{chen2024gazexplainlearningpredictnatural} & Full-data (W3D) [$\sim$70k samples] & 0.31 & 0.30 & 0.22 & 0.42 & 0.19 & 0.29 & 0.37 & \underline{0.55} & 0.17 & 0.20 & \underline{0.44} & 0.66 \\
LLada\textsuperscript{*}~\cite{zhou2025where} & Full-data (W3D) [$\sim$70k samples] & \textbf{0.44} & \underline{0.36} & \textbf{0.58} & \textbf{0.96} & \textbf{0.44} & \textbf{0.38} & \textbf{0.59} & \textbf{1.23} & \textbf{0.38} & \underline{0.32} & \textbf{0.52} & \textbf{1.00} \\
\midrule
\multicolumn{14}{l}{\textit{Zero-shot and In-Context Models}} \\
\midrule
Qwen-VL~\cite{bai2025qwen25vltechnicalreport} & Zero-shot (no training) 
& 0.10 & 0.19 & 0.28 & 0.34 & 0.19 
& 0.21 & 0.29 & 0.13 & 0.08 
& 0.21 & 0.29 & 0.12 \\

LLaVA~\cite{10.5555/3666122.3667638} & Zero-shot (no training) 
& 0.12 & 0.14 & 0.23 & 0.35 
& 0.13 & 0.19 & 0.11 & 0.10 
& 0.17 & 0.26 & 0.19 & 0.13 \\

Qwen-VL~\cite{bai2025qwen25vltechnicalreport} & In-context learning (no fine-tuning) 
& 0.13 & 0.18 & 0.22 & 0.36 & 0.21 
& 0.17 & 0.30 & 0.23 & 0.12 
& 0.24 & 0.33 & 0.15 \\

\midrule
\multicolumn{14}{l}{\textit{Few-shot Learning (BDD-A 90 samples)}} \\
\midrule
DeepGazeI\cite{kümmerer2015deepgazeiboosting} + LLaVA & Few-shot & 0.12 & 0.23 & 0.28 & 0.14 & 0.13 & 0.22 & 0.30 & 0.18 & 0.15 & 0.21 & 0.31 & 0.17 \\
DeepGazeIIE\cite{linardos2021deepgazeiiecalibratedprediction} + LLaVA & Few-shot & 0.11 & 0.18 & 0.26 & 0.17 & 0.11 & 0.20 & 0.32 & 0.13 & 0.11 & 0.19 & 0.34 & 0.10 \\
MLNet\cite{8356626} + LLaVA & Few-shot & 0.13 & 0.19 & 0.27 & 0.31 & 0.26 & 0.20 & 0.32 & 0.12 & 0.13 & 0.18 & 0.33 & 0.28 \\
\textbf{FSDAM (Ours)} & Few-shot & \underline{0.42} & \textbf{0.37} & \underline{0.48} & \underline{0.83} & \underline{0.35} & \underline{0.33} & \underline{0.46} & 0.47 & \underline{0.33} & \textbf{0.35} & 0.34 & \underline{0.84} \\
\bottomrule
\end{tabular}}
\label{tab:w3d_caption_comparison}
\end{table*}

\subsection{Qualitative Results}
\begin{figure}[t]
    \centering
    \includegraphics[width=1\linewidth]{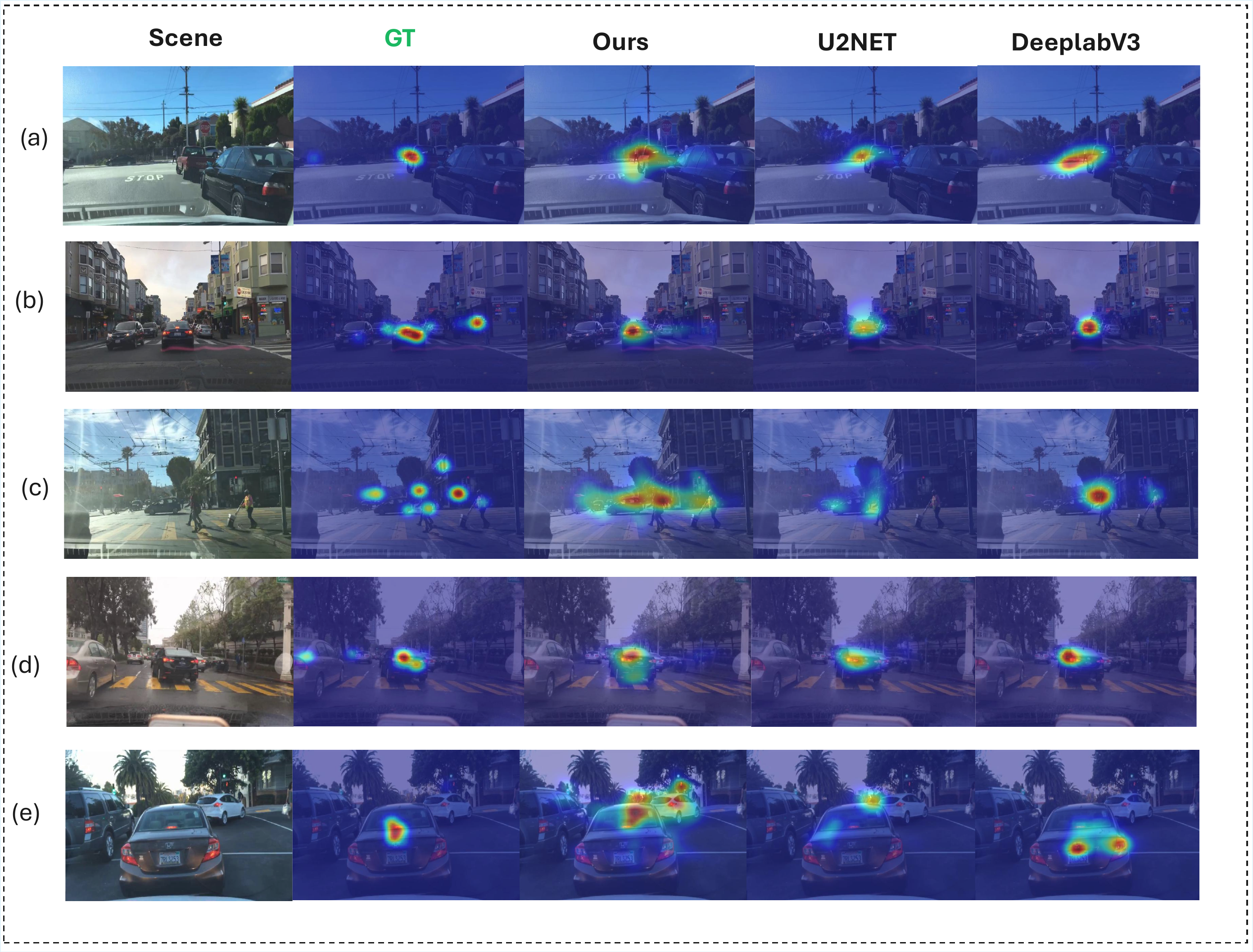}
    \caption{Qualitative comparison of driver attention prediction on BDD-A test scenes showing input image (left) and attention heatmaps from ground truth, FSDAM (90 samples), U2-Net\cite{Qin_2020_PR}, and DeepLabV3\cite{chen2017rethinkingatrousconvolutionsemantic}.}
    \label{fig:attn_pred}
\end{figure}

Figure~\ref{fig:attn_pred} compares FSDAM against fully-supervised U2-Net~\cite{Qin_2020_PR} and DeepLabV3~\cite{chen2017rethinkingatrousconvolutionsemantic}. In multi-agent scenarios (rows a, c, e), FSDAM produces broader attention coverage spanning peripheral pedestrians, intersection agents, and turning trajectories, whereas baselines concentrate on single targets or exhibit fragmented hot spots. In focused tasks (row d), all methods comparably localize attention to the lane-changing vehicle's brake lights. Row (b) reveals a shared limitation: all methods miss the distributed ground-truth pattern across both the lead vehicle and a right-side pedestrian. Overall, FSDAM achieves superior spatial coverage in complex scenes while maintaining competitive precision, validating that vision-language alignment enables effective few-shot attention learning.

\begin{figure}[t]
    \centering
    \begin{minipage}[t]{0.44\columnwidth}
        \centering
        \includegraphics[width=\linewidth, height=3.5cm, keepaspectratio]{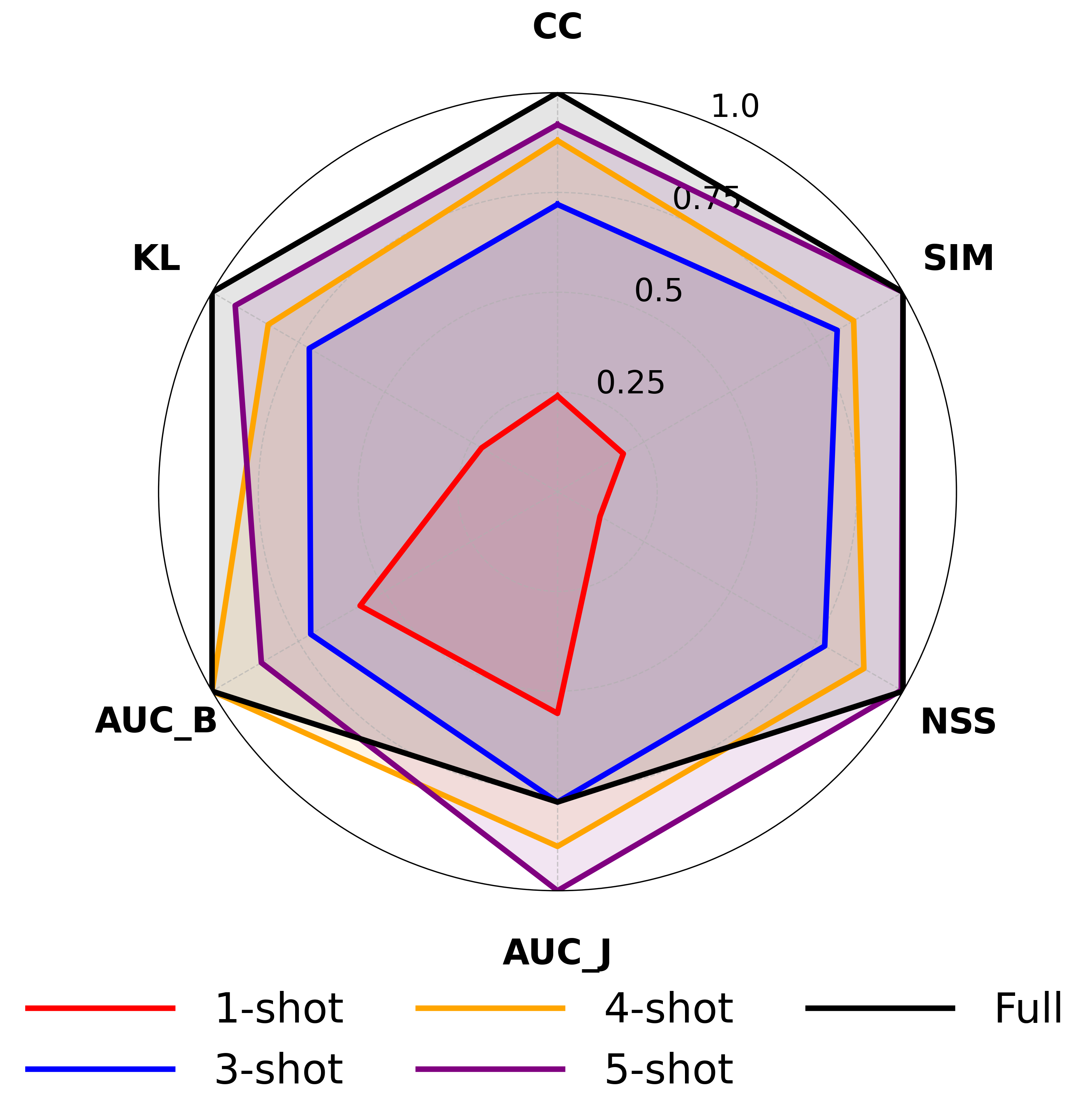}
        \caption{Few-shot learning performance on BDD-A. Metrics CC, SIM, NSS, AUC-J, AUC-B, and KL (inverted) are min-max normalized to [0,1]; larger area = better.}
        \label{fig:kshot_analysis}
    \end{minipage}
    \hfill
    \begin{minipage}[t]{0.52\columnwidth}
        \centering
        \includegraphics[width=\linewidth, height=3.5cm, keepaspectratio]{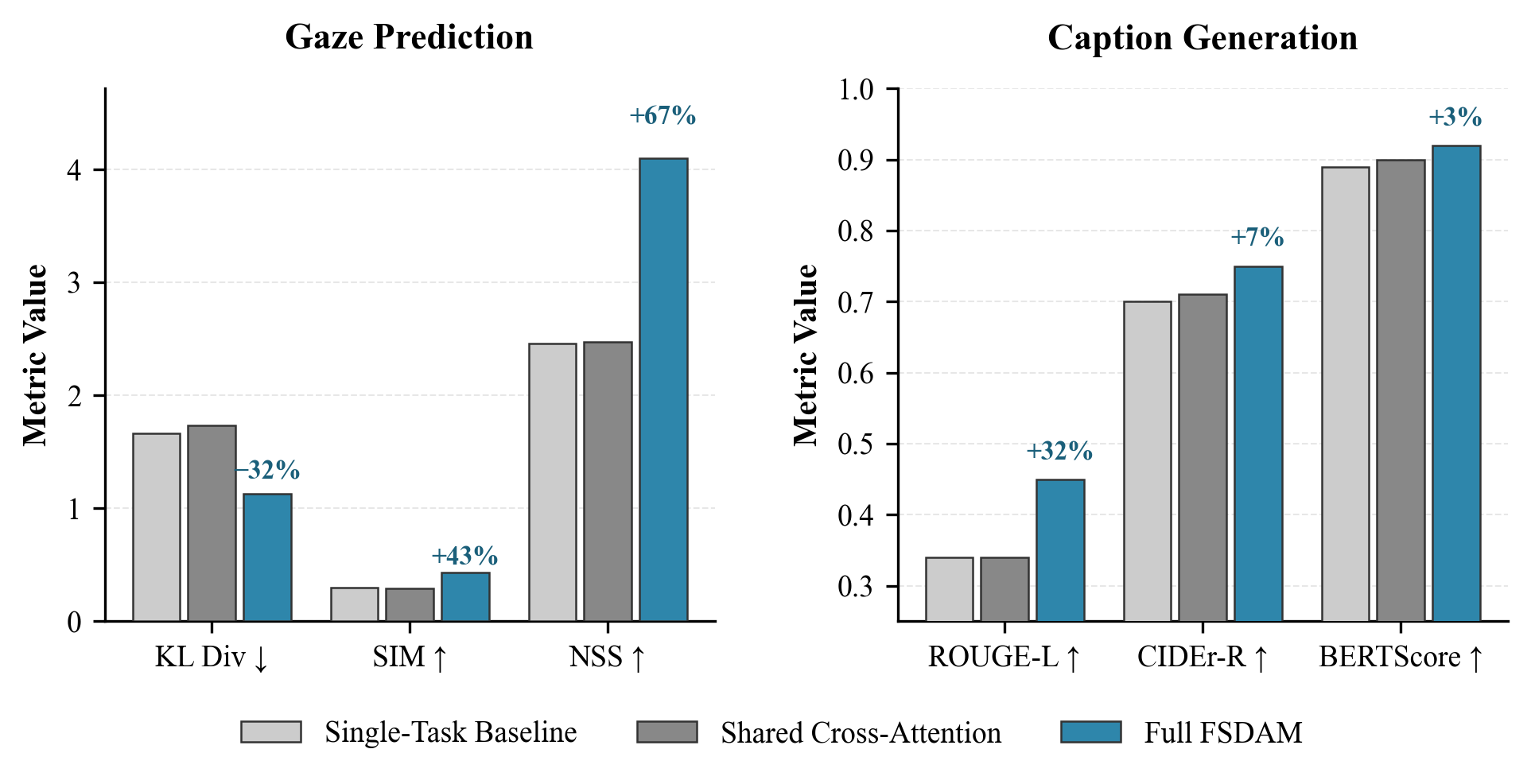}
        \caption{Ablation study comparing four architectural variants. Dual-pathway FSDAM achieves the largest gains for both gaze prediction and caption generation.}
        \label{fig:ablation}
    \end{minipage}
\end{figure}

\subsection{Few-Shot Learning Analysis}
Figure~\ref{fig:kshot_analysis} presents the impact of varying support set sizes on gaze prediction performance across six metrics. FSDAM demonstrates efficient learning dynamics, with substantial improvements from 1-shot to 5-shot settings. Performance gains are most pronounced between 1-shot and 3-shot (e.g., NSS increases from 2.32 to 3.64, CC improves from 0.41 to 0.53), after which improvements plateau. Notably, our 5-shot model achieves performance (CC 0.58, SIM 0.43, NSS 4.09, KL 1.17) that closely approaches the full-data baseline (CC 0.60, SIM 0.43, NSS 4.10, KL 1.13), recovering 96.7\% of full-data CC performance and 99.8\% of NSS performance with only five support samples. KL divergence shows a consistent reduction from 1.74 to 1.17 as support size increases, indicating improved distributional alignment between predicted and ground-truth gaze. Importantly, the fact that SIM saturates early (SIM 0.43 at 5-shot matches the full-data SIM 0.43) suggests that the model rapidly learns the overall attention shape, while additional shots primarily refine correlation and peak alignment reflected by CC/NSS. This trend is consistent with our design goal: training-only vision--language alignment provides a strong semantic prior, and the gaze pathway then sharpens spatial localization as supervision increases. Overall, meaningful gaze predictions emerge from minimal supervision, with performance largely saturating after a few examples.
\subsection{Ablations}
We conduct systematic ablation experiments to assess each component's contribution in FSDAM. We evaluate gaze prediction on the official BDD-A test set and caption generation on BDD-A curated samples with structured explanations, comprising all available test samples with complete four-component annotations. We compare four variants: \textbf{Gaze-Only} trains only gaze prediction; \textbf{Caption-Only} trains only caption generation; \textbf{Shared Cross-Attention} uses a single cross-attention module for both tasks; and \textbf{Full FSDAM} uses task-specific cross-attention modules with vision–language alignment. Human evaluation of the explanations is in the supplementary.

Figure~\ref{fig:ablation} shows a consistent performance hierarchy across both tasks. The shared cross-attention variant provides marginal gains over single-task baselines but degrades gaze prediction (4\% higher KL divergence), indicating negative transfer. In contrast, full FSDAM achieves substantial improvements: \textbf{31.9\% KL reduction}, \textbf{43.3\% SIM improvement}, and \textbf{66.7\% NSS improvement} for gaze prediction, alongside \textbf{32.4\% ROUGE-L}, \textbf{7.1\% CIDEr-R}, and \textbf{3.4\% BERTScore improvements} for caption generation, all relative to single-task baselines. This hierarchy (single-task $<$ shared $<$ dual-pathway) validates three design principles: (1) multi-task training can improve overall performance by leveraging shared visual representations; (2) forcing both tasks to share a single cross-attention module induces negative transfer (e.g., higher KL divergence); and (3) decoupled pathways with training-only vision--language alignment yield the strongest and most consistent gains across both tasks.
\section{Conclusions}
We present FSDAM, a framework that achieves joint gaze prediction and natural language explanation from 90 training examples. Through a dual-pathway architecture with training-only vision-language alignment, FSDAM achieves competitive performance against fully-supervised baselines across four benchmarks (BDD-A, DR(eye)VE, DADA-2000, W3D).
Ablation results confirm that task-specific cross-attention pathways prevent negative transfer, and that vision--language alignment provides meaningful semantic supervision to spatial prediction without increasing inference complexity. Few-shot learning analysis further shows that meaningful gaze predictions emerge from only a few examples, with performance approaching the full-data regime. Our results demonstrate the effectiveness of this approach across multiple benchmarks.
The model generates contextually grounded explanations across diverse driving scenarios, from normal conditions to safety-critical situations and accidents.
Limitations include dependence on a frozen CLIP backbone, whose features degrade under extreme weather, and single-frame inputs that cannot resolve temporal attention priority. Future work includes temporal modeling with video input and robustness to adverse conditions.

{
    \small
    \bibliographystyle{ieeenat_fullname}
    \bibliography{main}

\begin{thebibliography}{62}
\providecommand{\natexlab}[1]{#1}
\providecommand{\url}[1]{\texttt{#1}}
\expandafter\ifx\csname urlstyle\endcsname\relax
  \providecommand{\doi}[1]{doi: #1}\else
  \providecommand{\doi}{doi: \begingroup \urlstyle{rm}\Url}\fi

\bibitem[Alayrac et~al.(2022)Alayrac, Donahue, Luc, Miech, Barr, Hasson, Lenc, Mensch, Millican, Reynolds, Ring, Rutherford, Cabi, Han, Gong, Samangooei, Monteiro, Menick, Borgeaud, Brock, Nematzadeh, Sharifzadeh, Binkowski, Barreira, Vinyals, Zisserman, and Simonyan]{alayrac2022flamingo}
Jean-Baptiste Alayrac, Jeff Donahue, Pauline Luc, Antoine Miech, Iain Barr, Yana Hasson, Karel Lenc, Arthur Mensch, Katie Millican, Malcolm Reynolds, Roman Ring, Eliza Rutherford, Serkan Cabi, Tengda Han, Zhitao Gong, Sina Samangooei, Marianne Monteiro, Jacob Menick, Sebastian Borgeaud, Andrew Brock, Aida Nematzadeh, Sahand Sharifzadeh, Mikolaj Binkowski, Ricardo Barreira, Oriol Vinyals, Andrew Zisserman, and Karen Simonyan.
\newblock Flamingo: a visual language model for few-shot learning, 2022.

\bibitem[Alletto et~al.(2016)Alletto, Palazzi, Solera, Calderara, and Cucchiara]{7789504}
Stefano Alletto, Andrea Palazzi, Francesco Solera, Simone Calderara, and Rita Cucchiara.
\newblock Dr(eye)ve: A dataset for attention-based tasks with applications to autonomous and assisted driving.
\newblock In \emph{2016 IEEE Conference on Computer Vision and Pattern Recognition Workshops (CVPRW)}, pages 54--60, 2016.

\bibitem[Bai et~al.(2025)Bai, Chen, Liu, Wang, Ge, Song, Dang, Wang, Wang, Tang, Zhong, Zhu, Yang, Li, Wan, Wang, Ding, Fu, Xu, Ye, Zhang, Xie, Cheng, Zhang, Yang, Xu, and Lin]{bai2025qwen25vltechnicalreport}
Shuai Bai, Keqin Chen, Xuejing Liu, Jialin Wang, Wenbin Ge, Sibo Song, Kai Dang, Peng Wang, Shijie Wang, Jun Tang, Humen Zhong, Yuanzhi Zhu, Mingkun Yang, Zhaohai Li, Jianqiang Wan, Pengfei Wang, Wei Ding, Zheren Fu, Yiheng Xu, Jiabo Ye, Xi Zhang, Tianbao Xie, Zesen Cheng, Hang Zhang, Zhibo Yang, Haiyang Xu, and Junyang Lin.
\newblock Qwen2.5-vl technical report, 2025.

\bibitem[Bylinskii et~al.(2017)Bylinskii, Judd, Oliva, Torralba, and Durand]{bylinskii2017differentevaluationmetricstell}
Zoya Bylinskii, Tilke Judd, Aude Oliva, Antonio Torralba, and Frédo Durand.
\newblock What do different evaluation metrics tell us about saliency models?, 2017.

\bibitem[Carreira and Zisserman(2018)]{carreira2018quovadisactionrecognition}
Joao Carreira and Andrew Zisserman.
\newblock Quo vadis, action recognition? a new model and the kinetics dataset, 2018.

\bibitem[Chen et~al.(2017)Chen, Papandreou, Schroff, and Adam]{chen2017rethinkingatrousconvolutionsemantic}
Liang-Chieh Chen, George Papandreou, Florian Schroff, and Hartwig Adam.
\newblock Rethinking atrous convolution for semantic image segmentation, 2017.

\bibitem[Chen et~al.(2020)Chen, Kornblith, Norouzi, and Hinton]{chen2020simpleframeworkcontrastivelearning}
Ting Chen, Simon Kornblith, Mohammad Norouzi, and Geoffrey Hinton.
\newblock A simple framework for contrastive learning of visual representations, 2020.

\bibitem[Chen et~al.(2015)Chen, Fang, Lin, Vedantam, Gupta, Dollar, and Zitnick]{chen2015microsoftcococaptionsdata}
Xinlei Chen, Hao Fang, Tsung-Yi Lin, Ramakrishna Vedantam, Saurabh Gupta, Piotr Dollar, and C.~Lawrence Zitnick.
\newblock Microsoft coco captions: Data collection and evaluation server, 2015.

\bibitem[Chen et~al.(2024)Chen, Jiang, and Zhao]{chen2024gazexplainlearningpredictnatural}
Xianyu Chen, Ming Jiang, and Qi Zhao.
\newblock Gazexplain: Learning to predict natural language explanations of visual scanpaths, 2024.

\bibitem[Dodge and Karam(2018)]{8356626}
Samuel~F. Dodge and Lina~J. Karam.
\newblock Visual saliency prediction using a mixture of deep neural networks.
\newblock \emph{IEEE Transactions on Image Processing}, 27\penalty0 (8):\penalty0 4080--4090, 2018.

\bibitem[Dong et~al.(2024)Dong, Li, Dai, Zheng, Ma, Li, Xia, Xu, Wu, Liu, Chang, Sun, Li, and Sui]{dong2024surveyincontextlearning}
Qingxiu Dong, Lei Li, Damai Dai, Ce Zheng, Jingyuan Ma, Rui Li, Heming Xia, Jingjing Xu, Zhiyong Wu, Tianyu Liu, Baobao Chang, Xu Sun, Lei Li, and Zhifang Sui.
\newblock A survey on in-context learning, 2024.

\bibitem[Fan et~al.(2020)Fan, Zhuo, Tang, and Tai]{fan2020fewshotobjectdetectionattentionrpn}
Qi Fan, Wei Zhuo, Chi-Keung Tang, and Yu-Wing Tai.
\newblock Few-shot object detection with attention-rpn and multi-relation detector, 2020.

\bibitem[Fang et~al.(2019)Fang, Yan, Qiao, Xue, Wang, and Li]{fang2019dada2000drivingaccidentpredicted}
Jianwu Fang, Dingxin Yan, Jiahuan Qiao, Jianru Xue, He Wang, and Sen Li.
\newblock Dada-2000: Can driving accident be predicted by driver attention? analyzed by a benchmark, 2019.

\bibitem[Fang et~al.(2022)Fang, Yan, Qiao, Xue, and Yu]{9312486}
Jianwu Fang, Dingxin Yan, Jiahuan Qiao, Jianru Xue, and Hongkai Yu.
\newblock Dada: Driver attention prediction in driving accident scenarios.
\newblock \emph{IEEE Transactions on Intelligent Transportation Systems}, 23\penalty0 (6):\penalty0 4959--4971, 2022.

\bibitem[Fang et~al.(2023)Fang, Yan, Qiao, Xue, and Yu]{fang2023dadadriverattentionprediction}
Jianwu Fang, Dingxin Yan, Jiahuan Qiao, Jianru Xue, and Hongkai Yu.
\newblock Dada: Driver attention prediction in driving accident scenarios, 2023.

\bibitem[Fifty et~al.(2021)Fifty, Amid, Zhao, Yu, Anil, and Finn]{fifty2021efficientlyidentifyingtaskgroupings}
Christopher Fifty, Ehsan Amid, Zhe Zhao, Tianhe Yu, Rohan Anil, and Chelsea Finn.
\newblock Efficiently identifying task groupings for multi-task learning, 2021.

\bibitem[Fridman et~al.(2019)Fridman, Brown, Glazer, Angell, Dodd, Jenik, Terwilliger, Patsekin, Kindelsberger, Ding, Seaman, Mehler, Sipperley, Pettinato, Seppelt, Angell, Mehler, and Reimer]{8751968}
Lex Fridman, Daniel~E. Brown, Michael Glazer, William Angell, Spencer Dodd, Benedikt Jenik, Jack Terwilliger, Aleksandr Patsekin, Julia Kindelsberger, Li Ding, Sean Seaman, Alea Mehler, Andrew Sipperley, Anthony Pettinato, Bobbie~D. Seppelt, Linda Angell, Bruce Mehler, and Bryan Reimer.
\newblock Mit advanced vehicle technology study: Large-scale naturalistic driving study of driver behavior and interaction with automation.
\newblock \emph{IEEE Access}, 7:\penalty0 102021--102038, 2019.

\bibitem[Ghosh et~al.(2025)Ghosh, Zheng, Tamburo, Vuong, Alvarez-Padilla, Zhu, Cardei, Dunn, Mertz, and Narasimhan]{ghosh2025roadwork}
Anurag Ghosh, Shen Zheng, Robert Tamburo, Khiem Vuong, Juan Alvarez-Padilla, Hailiang Zhu, Michael Cardei, Nicholas Dunn, Christoph Mertz, and Srinivasa~G Narasimhan.
\newblock Roadwork: A dataset and benchmark for learning to recognize, observe, analyze and drive through work zones.
\newblock In \emph{ICCV}, 2025.

\bibitem[Grill et~al.(2020)Grill, Strub, Altché, Tallec, Richemond, Buchatskaya, Doersch, Pires, Guo, Azar, Piot, Kavukcuoglu, Munos, and Valko]{grill2020bootstrap}
Jean-Bastien Grill, Florian Strub, Florent Altché, Corentin Tallec, Pierre~H. Richemond, Elena Buchatskaya, Carl Doersch, Bernardo~Avila Pires, Zhaohan~Daniel Guo, Mohammad~Gheshlaghi Azar, Bilal Piot, Koray Kavukcuoglu, Rémi Munos, and Michal Valko.
\newblock Bootstrap your own latent: A new approach to self-supervised learning, 2020.

\bibitem[Houlsby et~al.(2019)Houlsby, Giurgiu, Jastrzebski, Morrone, de~Laroussilhe, Gesmundo, Attariyan, and Gelly]{houlsby2019parameterefficienttransferlearningnlp}
Neil Houlsby, Andrei Giurgiu, Stanislaw Jastrzebski, Bruna Morrone, Quentin de Laroussilhe, Andrea Gesmundo, Mona Attariyan, and Sylvain Gelly.
\newblock Parameter-efficient transfer learning for nlp, 2019.

\bibitem[Hu et~al.(2022)Hu, Shen, Wallis, Allen-Zhu, Li, Wang, Wang, and Chen]{hu2022lora}
Edward~J Hu, Yelong Shen, Phillip Wallis, Zeyuan Allen-Zhu, Yuanzhi Li, Shean Wang, Lu Wang, and Weizhu Chen.
\newblock Lo{RA}: Low-rank adaptation of large language models.
\newblock In \emph{International Conference on Learning Representations}, 2022.

\bibitem[Jiang et~al.(2023)Jiang, Sablayrolles, Mensch, Bamford, Chaplot, de~las Casas, Bressand, Lengyel, Lample, Saulnier, Lavaud, Lachaux, Stock, Scao, Lavril, Wang, Lacroix, and Sayed]{jiang2023mistral7b}
Albert~Q. Jiang, Alexandre Sablayrolles, Arthur Mensch, Chris Bamford, Devendra~Singh Chaplot, Diego de~las Casas, Florian Bressand, Gianna Lengyel, Guillaume Lample, Lucile Saulnier, Lélio~Renard Lavaud, Marie-Anne Lachaux, Pierre Stock, Teven~Le Scao, Thibaut Lavril, Thomas Wang, Timothée Lacroix, and William~El Sayed.
\newblock Mistral 7b, 2023.

\bibitem[Kang et~al.(2019)Kang, Liu, Wang, Yu, Feng, and Darrell]{kang2019fewshotobjectdetectionfeature}
Bingyi Kang, Zhuang Liu, Xin Wang, Fisher Yu, Jiashi Feng, and Trevor Darrell.
\newblock Few-shot object detection via feature reweighting, 2019.

\bibitem[Kim et~al.(2023)Kim, Kim, Cho, Luo, and Hong]{kim2023universalfewshotlearningdense}
Donggyun Kim, Jinwoo Kim, Seongwoong Cho, Chong Luo, and Seunghoon Hong.
\newblock Universal few-shot learning of dense prediction tasks with visual token matching, 2023.

\bibitem[Kotseruba and Tsotsos(2024)]{10588743}
Iuliia Kotseruba and John~K. Tsotsos.
\newblock Scout+: Towards practical task-driven drivers’ gaze prediction.
\newblock In \emph{2024 IEEE Intelligent Vehicles Symposium (IV)}, pages 1927--1932, 2024.

\bibitem[Kr{\"o}ger et~al.(2020)Kr{\"o}ger, Lutz, and M{\"u}ller]{kroger2020does}
Jacob~Leon Kr{\"o}ger, Otto Hans-Martin Lutz, and Florian M{\"u}ller.
\newblock What does your gaze reveal about you? on the privacy implications of eye tracking.
\newblock In \emph{IFIP International Summer School on Privacy and Identity Management}, pages 226--241. Springer, 2020.

\bibitem[Kümmerer et~al.(2015)Kümmerer, Theis, and Bethge]{kümmerer2015deepgazeiboosting}
Matthias Kümmerer, Lucas Theis, and Matthias Bethge.
\newblock Deep gaze i: Boosting saliency prediction with feature maps trained on imagenet, 2015.

\bibitem[Kümmerer et~al.(2016)Kümmerer, Wallis, and Bethge]{kümmerer2016deepgazeiireadingfixations}
Matthias Kümmerer, Thomas S.~A. Wallis, and Matthias Bethge.
\newblock Deepgaze ii: Reading fixations from deep features trained on object recognition, 2016.

\bibitem[Li et~al.(2024)Li, Sharma, and Zhang]{Li_2024}
Dongyue Li, Aneesh Sharma, and Hongyang~R. Zhang.
\newblock Scalable multitask learning using gradient-based estimation of task affinity.
\newblock In \emph{Proceedings of the 30th ACM SIGKDD Conference on Knowledge Discovery and Data Mining}, page 1542–1553. ACM, 2024.

\bibitem[Li et~al.(2019)Li, Wang, Zhu, Sui, Wang, Qu, and Green]{li2019drivers}
Guofa Li, Ying Wang, Fangping Zhu, Xiaoxuan Sui, Ning Wang, Xingda Qu, and Paul Green.
\newblock Drivers’ visual scanning behavior at signalized and unsignalized intersections: A naturalistic driving study in china.
\newblock \emph{Journal of safety research}, 71:\penalty0 219--229, 2019.

\bibitem[Li et~al.(2022)Li, Li, Xiong, and Hoi]{li2022blip}
Junnan Li, Dongxu Li, Caiming Xiong, and Steven Hoi.
\newblock Blip: Bootstrapping language-image pre-training for unified vision-language understanding and generation.
\newblock In \emph{ICML}, 2022.

\bibitem[Li and Liang(2021)]{li2021prefixtuning}
Xiang~Lisa Li and Percy Liang.
\newblock Prefix-tuning: Optimizing continuous prompts for generation, 2021.

\bibitem[Li et~al.(2025)Li, Xiong, Guo, Li, Yan, Xu, Zhou, Chen, Sun, Wang, et~al.]{li2025recogdrive}
Yongkang Li, Kaixin Xiong, Xiangyu Guo, Fang Li, Sixu Yan, Gangwei Xu, Lijun Zhou, Long Chen, Haiyang Sun, Bing Wang, et~al.
\newblock Recogdrive: A reinforced cognitive framework for end-to-end autonomous driving.
\newblock \emph{arXiv preprint arXiv:2506.08052}, 2025.

\bibitem[Linardos et~al.(2021)Linardos, Kümmerer, Press, and Bethge]{linardos2021deepgazeiiecalibratedprediction}
Akis Linardos, Matthias Kümmerer, Ori Press, and Matthias Bethge.
\newblock Deepgaze iie: Calibrated prediction in and out-of-domain for state-of-the-art saliency modeling, 2021.

\bibitem[Liu et~al.(2023)Liu, Li, Wu, and Lee]{10.5555/3666122.3667638}
Haotian Liu, Chunyuan Li, Qingyang Wu, and Yong~Jae Lee.
\newblock Visual instruction tuning.
\newblock In \emph{Proceedings of the 37th International Conference on Neural Information Processing Systems}, Red Hook, NY, USA, 2023. Curran Associates Inc.

\bibitem[Liu et~al.(2024)Liu, Li, Li, Li, Zhang, Shen, and Lee]{liu2024llavanext}
Haotian Liu, Chunyuan Li, Yuheng Li, Bo Li, Yuanhan Zhang, Sheng Shen, and Yong~Jae Lee.
\newblock Llava-next: Improved reasoning, ocr, and world knowledge, 2024.

\bibitem[Long et~al.(2015)Long, Shelhamer, and Darrell]{long2015fullyconvolutionalnetworkssemantic}
Jonathan Long, Evan Shelhamer, and Trevor Darrell.
\newblock Fully convolutional networks for semantic segmentation, 2015.

\bibitem[Malla et~al.(2023)Malla, Choi, Dwivedi, Choi, and Li]{malla2023drama}
Srikanth Malla, Chiho Choi, Isht Dwivedi, Joon~Hee Choi, and Jiachen Li.
\newblock Drama: Joint risk localization and captioning in driving.
\newblock In \emph{Proceedings of the IEEE/CVF Winter Conference on Applications of Computer Vision}, pages 1043--1052, 2023.

\bibitem[Mori et~al.(2019)Mori, Fukui, Hirakawa, Nishiyama, Yamashita, and Fujiyoshi]{8917187}
Yuki Mori, Hiroshi Fukui, Tsubasa Hirakawa, Jo Nishiyama, Takayoshi Yamashita, and Hironobu Fujiyoshi.
\newblock Attention neural baby talk: Captioning of risk factors while driving.
\newblock In \emph{2019 IEEE Intelligent Transportation Systems Conference (ITSC)}, pages 4317--4322, 2019.

\bibitem[OpenAI(2024)]{openai2024gpt4technicalreport}
OpenAI.
\newblock Gpt-4 technical report, 2024.

\bibitem[Palazzi et~al.(2018)Palazzi, Abati, Solera, and Cucchiara]{palazzi2018predicting}
Andrea Palazzi, Davide Abati, Francesco Solera, and Rita Cucchiara.
\newblock Predicting the driver's focus of attention: the dr (eye) ve project.
\newblock \emph{IEEE transactions on pattern analysis and machine intelligence}, 41\penalty0 (7):\penalty0 1720--1733, 2018.

\bibitem[Pang et~al.(2020)Pang, Zhao, Zhang, and Lu]{pang2020multiscaleinteractivenetworksalient}
Youwei Pang, Xiaoqi Zhao, Lihe Zhang, and Huchuan Lu.
\newblock Multi-scale interactive network for salient object detection, 2020.

\bibitem[Paszke et~al.(2019)Paszke, Gross, Massa, Lerer, Bradbury, Chanan, Killeen, Lin, Gimelshein, Antiga, Desmaison, Köpf, Yang, DeVito, Raison, Tejani, Chilamkurthy, Steiner, Fang, Bai, and Chintala]{paszke2019pytorchimperativestylehighperformance}
Adam Paszke, Sam Gross, Francisco Massa, Adam Lerer, James Bradbury, Gregory Chanan, Trevor Killeen, Zeming Lin, Natalia Gimelshein, Luca Antiga, Alban Desmaison, Andreas Köpf, Edward Yang, Zach DeVito, Martin Raison, Alykhan Tejani, Sasank Chilamkurthy, Benoit Steiner, Lu Fang, Junjie Bai, and Soumith Chintala.
\newblock Pytorch: An imperative style, high-performance deep learning library, 2019.

\bibitem[Qin et~al.(2020)Qin, Zhang, Huang, Dehghan, Zaiane, and Jagersand]{Qin_2020_PR}
Xuebin Qin, Zichen Zhang, Chenyang Huang, Masood Dehghan, Osmar Zaiane, and Martin Jagersand.
\newblock U2-net: Going deeper with nested u-structure for salient object detection.
\newblock \emph{Pattern Recognition}, 106:\penalty0 107404, 2020.

\bibitem[Radford et~al.(2021)Radford, Kim, Hallacy, Ramesh, Goh, Agarwal, Sastry, Askell, Mishkin, Clark, Krueger, and Sutskever]{radford2021learning}
Alec Radford, Jong~Wook Kim, Chris Hallacy, Aditya Ramesh, Gabriel Goh, Sandhini Agarwal, Girish Sastry, Amanda Askell, Pamela Mishkin, Jack Clark, Gretchen Krueger, and Ilya Sutskever.
\newblock Learning transferable visual models from natural language supervision, 2021.

\bibitem[Ronneberger et~al.(2015)Ronneberger, Fischer, and Brox]{ronneberger2015unetconvolutionalnetworksbiomedical}
Olaf Ronneberger, Philipp Fischer, and Thomas Brox.
\newblock U-net: Convolutional networks for biomedical image segmentation, 2015.

\bibitem[Ruiz et~al.(2023)Ruiz, Li, Jampani, Pritch, Rubinstein, and Aberman]{ruiz2023dreamboothfinetuningtexttoimage}
Nataniel Ruiz, Yuanzhen Li, Varun Jampani, Yael Pritch, Michael Rubinstein, and Kfir Aberman.
\newblock Dreambooth: Fine tuning text-to-image diffusion models for subject-driven generation, 2023.

\bibitem[Sharma and Chakraborty(2024)]{sharma2024review}
Pavan~Kumar Sharma and Pranamesh Chakraborty.
\newblock A review of driver gaze estimation and application in gaze behavior understanding.
\newblock \emph{Engineering Applications of Artificial Intelligence}, 133:\penalty0 108117, 2024.

\bibitem[Shlens(2014)]{shlens2014noteskullbackleiblerdivergencelikelihood}
Jonathon Shlens.
\newblock Notes on kullback-leibler divergence and likelihood, 2014.

\bibitem[Sima et~al.(2023)Sima, Renz, Chitta, Chen, Zhang, Xie, Beißwenger, Luo, Geiger, and Li]{sima2023drivelm}
Chonghao Sima, Katrin Renz, Kashyap Chitta, Li Chen, Hanxue Zhang, Chengen Xie, Jens Beißwenger, Ping Luo, Andreas Geiger, and Hongyang Li.
\newblock Drivelm: Driving with graph visual question answering, 2023.

\bibitem[Steelman et~al.(2017)Steelman, McCarley, and Wickens]{steelman2017theory}
Kelly~S Steelman, Jason~S McCarley, and Christopher~D Wickens.
\newblock Theory-based models of attention in visual workspaces.
\newblock \emph{International Journal of Human--Computer Interaction}, 33\penalty0 (1):\penalty0 35--43, 2017.

\bibitem[Tian et~al.(2022)Tian, Deng, and Yan]{9812524}
Han Tian, Tao Deng, and Hongmei Yan.
\newblock Driving as well as on a sunny day? predicting driver's fixation in rainy weather conditions via a dual-branch visual model.
\newblock \emph{IEEE/CAA Journal of Automatica Sinica}, 9\penalty0 (7):\penalty0 1335--1338, 2022.

\bibitem[Tian et~al.(2024)Tian, Gu, Li, Liu, Zhao, Wang, Zhan, Jia, Lang, and Zhao]{DriveVLM}
Xiaoyu Tian, Junru Gu, Bailin Li, Yicheng Liu, Zhiyong Zhao, Yang Wang, Kun Zhan, Peng Jia, Xianpeng Lang, and Hang Zhao.
\newblock Drivevlm: The convergence of autonomous driving and large vision-language models.
\newblock \emph{arXiv preprint arXiv:2402.12289}, 2024.

\bibitem[van~den Oord et~al.(2018)van~den Oord, Li, and Vinyals]{oord2018representation}
Aaron van~den Oord, Yazhe Li, and Oriol Vinyals.
\newblock Representation learning with contrastive predictive coding, 2018.

\bibitem[Vaswani et~al.(2017)Vaswani, Shazeer, Parmar, Uszkoreit, Jones, Gomez, Kaiser, and Polosukhin]{vaswani2017attention}
Ashish Vaswani, Noam Shazeer, Niki Parmar, Jakob Uszkoreit, Llion Jones, Aidan~N. Gomez, Lukasz Kaiser, and Illia Polosukhin.
\newblock Attention is all you need, 2017.

\bibitem[Wang et~al.(2020)Wang, Liew, Zou, Zhou, and Feng]{wang2020panetfewshotimagesemantic}
Kaixin Wang, Jun~Hao Liew, Yingtian Zou, Daquan Zhou, and Jiashi Feng.
\newblock Panet: Few-shot image semantic segmentation with prototype alignment, 2020.

\bibitem[Xia et~al.(2018)Xia, Zhang, Kim, Nakayama, Zipser, and Whitney]{xia2018predictingdriverattentioncritical}
Ye Xia, Danqing Zhang, Jinkyu Kim, Ken Nakayama, Karl Zipser, and David Whitney.
\newblock Predicting driver attention in critical situations, 2018.

\bibitem[Xia et~al.(2019)Xia, Zhang, Kim, Nakayama, Zipser, and Whitney]{10.1007/978-3-030-20873-8_42}
Ye Xia, Danqing Zhang, Jinkyu Kim, Ken Nakayama, Karl Zipser, and David Whitney.
\newblock Predicting driver attention in critical situations.
\newblock In \emph{Computer Vision -- ACCV 2018}, pages 658--674, Cham, 2019. Springer International Publishing.

\bibitem[Xu et~al.(2024)Xu, Zhang, Xie, Zhao, Guo, Wong, Li, and Zhao]{xu2024drivegpt4interpretableendtoendautonomous}
Zhenhua Xu, Yujia Zhang, Enze Xie, Zhen Zhao, Yong Guo, Kwan-Yee.~K. Wong, Zhenguo Li, and Hengshuang Zhao.
\newblock Drivegpt4: Interpretable end-to-end autonomous driving via large language model, 2024.

\bibitem[Yan et~al.(2024)Yan, Fang, Lv, and Su]{yan2024anomalysdfewshotmulticlassanomaly}
Zhenyu Yan, Qingqing Fang, Wenxi Lv, and Qinliang Su.
\newblock Anomalysd: Few-shot multi-class anomaly detection with stable diffusion model, 2024.

\bibitem[Zhou et~al.(2024)Zhou, Liu, Yurtsever, Zagar, Zimmer, Cao, and Knoll]{zhou2024visionlanguagemodelsautonomous}
Xingcheng Zhou, Mingyu Liu, Ekim Yurtsever, Bare~Luka Zagar, Walter Zimmer, Hu Cao, and Alois~C. Knoll.
\newblock Vision language models in autonomous driving: A survey and outlook, 2024.

\bibitem[Zhou et~al.(2025)Zhou, Tang, Xiao, Lin, Liu, Guo, Fei, Xia, and Gou]{zhou2025where}
Yuchen Zhou, Jiayu Tang, Xiaoyan Xiao, Yueyao Lin, Linkai Liu, Zipeng Guo, Hao Fei, Xiaobo Xia, and Chao Gou.
\newblock Where, what, why: Towards explainable driver attention prediction.
\newblock In \emph{Proceedings of the IEEE/CVF International Conference on Computer Vision (ICCV)}, 2025.

\end{thebibliography}
}
\clearpage
\setcounter{page}{1}
\maketitlesupplementary

\section{Few-Shot Data Curation Strategy}
\label{sec:data_curation}
\begin{figure}[t]
\centering
\includegraphics[width=\linewidth]{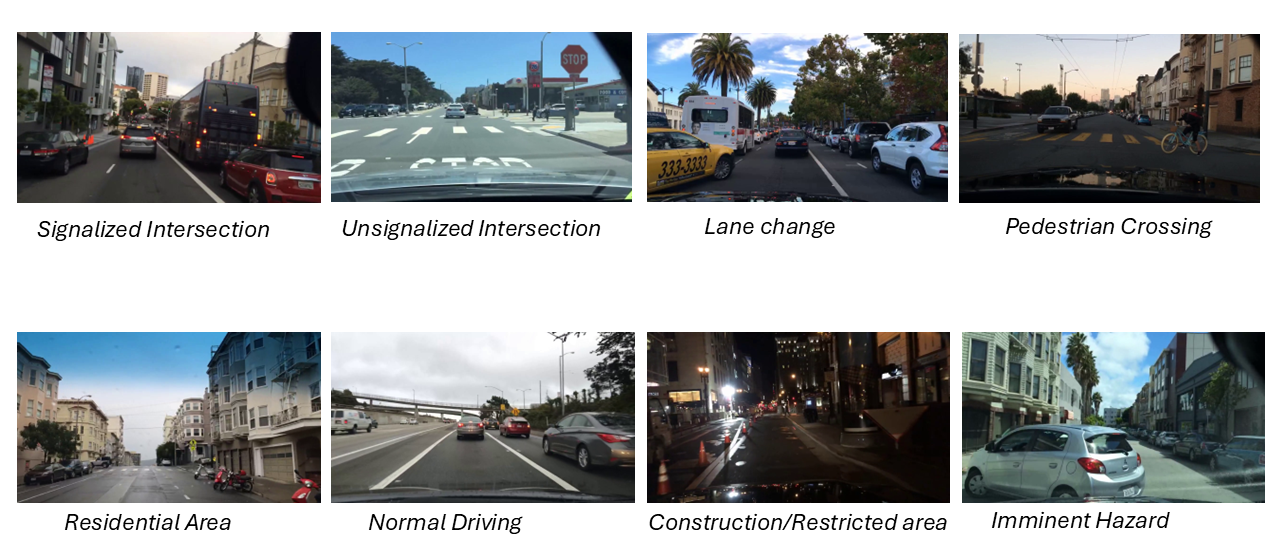}
\caption{\textbf{Few-shot training scenarios.} Our curated dataset covers eight critical driving contexts: signalized intersections, unsignalized intersections, lane changes, pedestrian crossings, residential areas, normal highway driving, construction zones, and imminent hazards to capture distinct gaze patterns essential for safe navigation \cite{li2019drivers, sharma2024review}.}
\label{fig:teaser}
\end{figure}
\subsection{Overview}

We present a principled approach to curating few-shot training data that capture the temporal dynamics of driver attention. We develop a few-shot modeling framework that trains a general-purpose vision–language model to predict and explain the temporal dynamics of driver attention across diverse scenarios. Our approach identifies critical gaze transitions and extracts frame pairs that encode meaningful attention shifts in safety-critical scenarios across diverse driving scenarios. In this paper, we instantiate the framework on the BDD-A dataset, which consists of dash-cam video clips centered on driver-initiated braking events. From BDD-A, we curate 90 high-quality training samples spanning eight types of safety-critical driving contexts to demonstrate how human drivers reallocate attention in such scenarios. (Few examples are given in Figure~\ref{fig:teaser}).


\subsection{Temporal Gaze Dynamics Mining}

Our curation pipeline applies an information-theoretic analysis to the temporal structure of gaze patterns. Specifically, we compute the Kullback–Leibler (KL) divergence~\cite{shlens2014noteskullbackleiblerdivergencelikelihood} between consecutive gaze maps, using KL peaks to pin-point moments of maximal change in the attention distribution. We treat these high-divergence moments as critical attention shifts and select frame pairs around them as the most informative training examples.

\noindent\textbf{Parameter Selection.} 
Algorithm~\ref{alg:mining} operates with four parameters, each motivated by properties of the BDD-A data and the temporal structure of 
driver gaze. $K{=}2$ bounds each clip's contribution to at most two pairs, preventing any single video from dominating the training
distribution while maintaining temporal diversity, following temporal sampling strategies from video action 
recognition~\cite{carreira2018quovadisactionrecognition}. $\Delta_{\min}{=}3$ and $\Delta_{\max}{=}18$ define the 
temporal search window at 30fps, corresponding to 100--600ms --- the range within which a driver fixation 
transition is perceptually meaningful and attributable to a single attention shift. $r{=}25$ enforces minimum 
anchor spacing to guarantee independence between selected pairs within the same clip. Applied to all 
1000 BDD-A clips, after filtering clips shorter than 50 frames following~\cite{carreira2018quovadisactionrecognition}, 
the algorithm yielded 651 candidate pairs; the remaining clips were excluded due to insufficient gaze dynamics or failure to satisfy the $\Delta_{\min}$ constraint within the available temporal window.

\begin{algorithm}[t]
\caption{Temporal Gaze Transition Mining}
\label{alg:mining}
\begin{algorithmic}[1]
\Require Synchronized gaze video $G$ and RGB video $V$ with $T$ frames
\Require Parameters $K=2$, $\Delta_{\min}=3$, $\Delta_{\max}=18$, $r=25$
\Ensure Frame pairs that maximize gaze transitions

\For{$t = 1$ to $T$}
    \State $h_t \gets \mathrm{hist}(G_t)$
\EndFor

\For{$t = 2$ to $T$}
    \State $s_t \gets \mathrm{KL}(h_t \,\|\, h_{t-1})$
\EndFor

\State $\bar{s} \gets \mathrm{smooth}(s, 5)$
\State $P \gets \{\, t \mid \bar{s}_t > \bar{s}_{t-1} \;\land\; \bar{s}_t > \bar{s}_{t+1} \,\}$
\State $A \gets \mathrm{NMS}(P, \bar{s}, r)$ \Comment{Enforce spacing}
\State $\mathcal{C} \gets \emptyset$

\For{each anchor $t \in A$}
    \State $\delta_{\max} \gets \min(\Delta_{\max}, T-t)$
    \If{$\delta_{\max} \geq \Delta_{\min}$}
        \State $\delta_{\mathrm{opt}} \gets \arg\max\limits_{\delta \in [\Delta_{\min}, \delta_{\max}]}
        \mathrm{KL}(h_{t+\delta} \,\|\, h_t)$
        \State $\mathcal{C} \gets \mathcal{C} \cup \{(t,\, t+\delta_{\mathrm{opt}},\, \mathrm{KL}(h_{t+\delta_{\mathrm{opt}}} \,\|\, h_t))\}$
    \EndIf
\EndFor

\State Sort $\mathcal{C}$ by KL score in descending order
\State Select top $K$ pairs with mutual spacing at least $r$ frames
\State Export $\{(V_t, G_t, V_{t+\delta}, G_{t+\delta})\}$ for the selected pairs
\end{algorithmic}
\end{algorithm}

\subsection{Structured Annotation Protocol}
\label{sec:annotation}
\begin{figure*}[t]
\centering
\includegraphics[width=\textwidth]{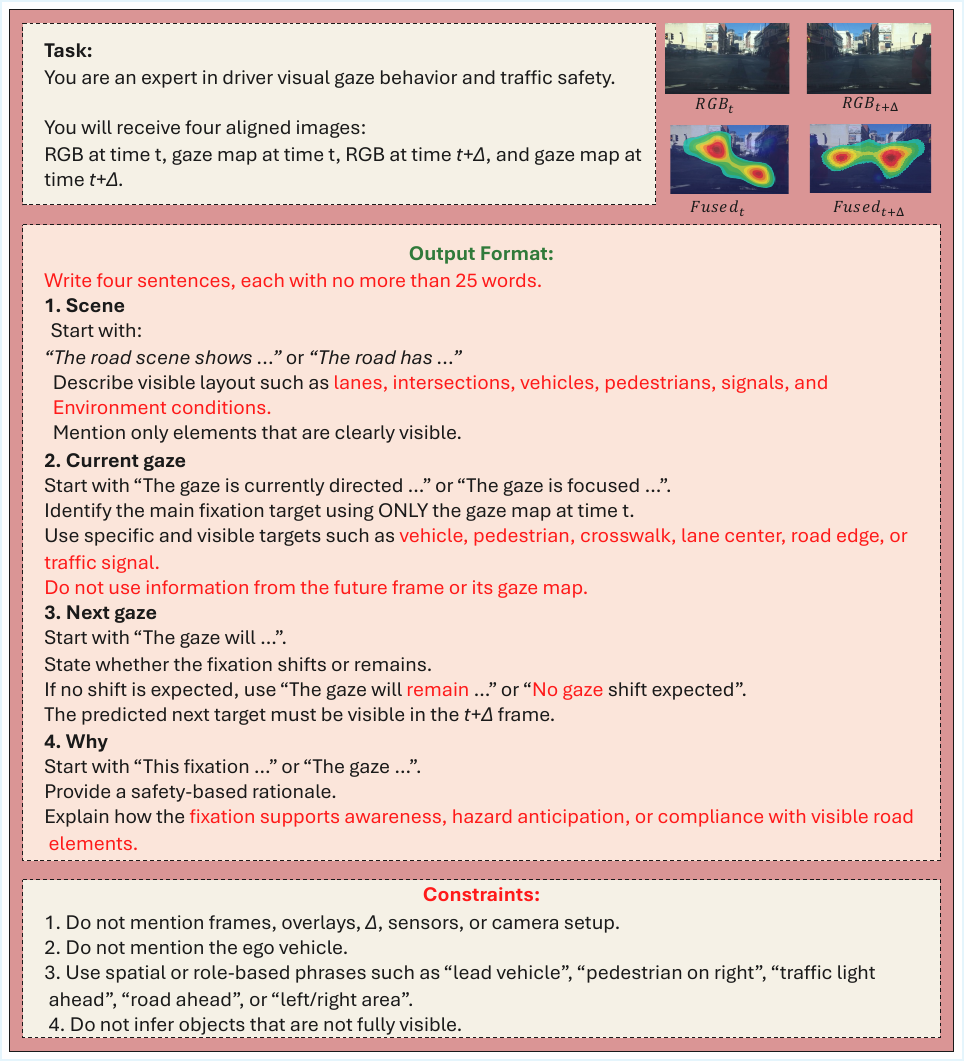}
\caption{\textbf{Structured annotation framework.} Our four-sentence protocol ensures consistent, high-quality descriptions across all training pairs. Each sentence serves a specific purpose: (1) scene context establishment, (2) current attention state, (3) future attention prediction, and (4) safety-driven rationale. This structured approach enables the model to learn the causal relationship between scene dynamics and attention shifts.}
\label{fig:annotation_prompt}
\end{figure*}

To generate consistent and informative captions, we developed a structured annotation protocol (Figure~\ref{fig:annotation_prompt}) that decomposes each gaze transition into four components:

\begin{itemize}
    \item \textbf{Scene Context:} Establishes the environmental layout and traffic elements visible in the frame, providing spatial grounding for attention prediction.
    
    \item \textbf{Current gaze:} Identifies the primary attention target at time $t$ based solely on the gaze heatmap, ensuring that annotations are grounded in actual gaze data.
    
    \item \textbf{Next Gaze:} Predicts the shift of attention to time $t+\Delta$, capturing the temporal evolution of gaze patterns.
    
    \item \textbf{Safety Rationale (Why):} Explains the driving-relevant motivation behind the attention transition, linking gaze behavior to safe navigation principles.
\end{itemize}

Each component is constrained to 25 words, enforcing conciseness while maintaining descriptive clarity. This rigid structure ensures consistency and preciseness across annotations while capturing the causal relationships essential for few-shot learning. The constraints encourage the model to capture fine-grained gaze differences across scenarios while enforcing a fixed output length, which helps reduce length-induced data bias.

\subsection{Quality Control and Refinement}

Our annotation pipeline combines automated caption initialization with human refinement. GPT-4o was used only to initialize structured captions from the selected RGB frame pair, gaze heatmaps, and the fixed four-component prompt. The generated captions were then independently reviewed by two expert annotators to ensure spatial accuracy, temporal consistency, format compliance, and driving relevance.

The annotators verified four criteria. First, \textit{pixel grounding}: each caption must refer only to visible scene elements and gaze-supported regions. Second, \textit{format compliance}: each annotation must follow the four-component structure consisting of scene context, current gaze, next gaze, and safety rationale. Third, \textit{temporal consistency}: the current-gaze sentence must describe attention at time $t$, while the next-gaze sentence must describe the short-horizon target at time $t+\Delta$ without using information beyond the paired future frame. Fourth, \textit{causal correctness}: the safety rationale must identify a concrete driving-relevant visual cue, such as a crosswalk, brake light, merging vehicle, pedestrian movement, or right-of-way conflict.

Across the 90 curated samples, 52 captions were accepted without modification, 28 required minor edits, 8 required major edits, and 2 were regenerated. Common corrections included improving object specificity, fixing gaze-target mismatch, resolving temporal inconsistency, removing hallucinated objects, and replacing generic rationales with concrete visual cues. This verification step ensures that the final annotations are not raw language-model outputs, but human-validated structured supervision grounded in both the visual scene and the corresponding gaze transition.

\subsection{Dataset Statistics and Distribution}
\begin{table}[t]
\centering
\caption{Few-shot dataset composition across driving scenarios}
\label{tab:dataset_stats}
\begin{tabular}{lcc}
\toprule
\textbf{Scenario Type} & \textbf{Frame Pairs} & \textbf{Percentage} \\
\midrule
Signalized Intersection & 14 & 15.6\% \\
Unsignalized Intersection & 12 & 13.3\% \\
Lane Change & 13 & 14.4\% \\
Pedestrian Crossing & 11 & 12.2\% \\
Residential Area & 10 & 11.1\% \\
Normal Driving & 12 & 13.3\% \\
Construction Zone & 9 & 10.0\% \\
Imminent Hazard & 9 & 10.0\% \\
\midrule
\textbf{Total} & \textbf{90} & \textbf{100\%} \\
\bottomrule
\end{tabular}
\end{table}

The 651 candidate pairs were categorized into eight scenario types (Figure \ref{fig:teaser}), unevenly distributed across categories, reflecting the natural frequency of each scenario type in BDD-A. To mitigate potential bias from category imbalance while constructing a slimmer few-shot training set, we applied stratified sampling to the candidate pairs. Specifically,  within each of the eight scenario categories, pairs were ranked by KL divergence score and the top 30  pairs from each category were retained, following prior few-shot studies in semantic segmentation and object detection\cite{kang2019fewshotobjectdetectionfeature,fan2020fewshotobjectdetectionattentionrpn,yan2024anomalysdfewshotmulticlassanomaly}. We further reduce the candidate pool to  9--14 pairs per category to minimize redundancy while preserving diversity in weather, lighting and scenario characteristics. This process resulted in 90 frame pairs with balanced representation across all eight scenario categories (Table~\ref{tab:dataset_stats}). Each pair is then annotated with a structured caption following the protocol described in Section~\ref{sec:annotation}, yielding one training sample per pair --- capturing a complete attention transition from initial fixation through gaze shift to new target acquisition --- 
providing rich supervision for few-shot learning.

\subsection{Reproducibility}
The project repository is available at \url{https://github.com/fsdam-vlc/fsdam}.

\section{Additional Qualitative Analysis}

\subsection{More examples on BDD-A Test Set}
\begin{figure*}[t]
    \centering
    \includegraphics[width=0.8\textwidth]{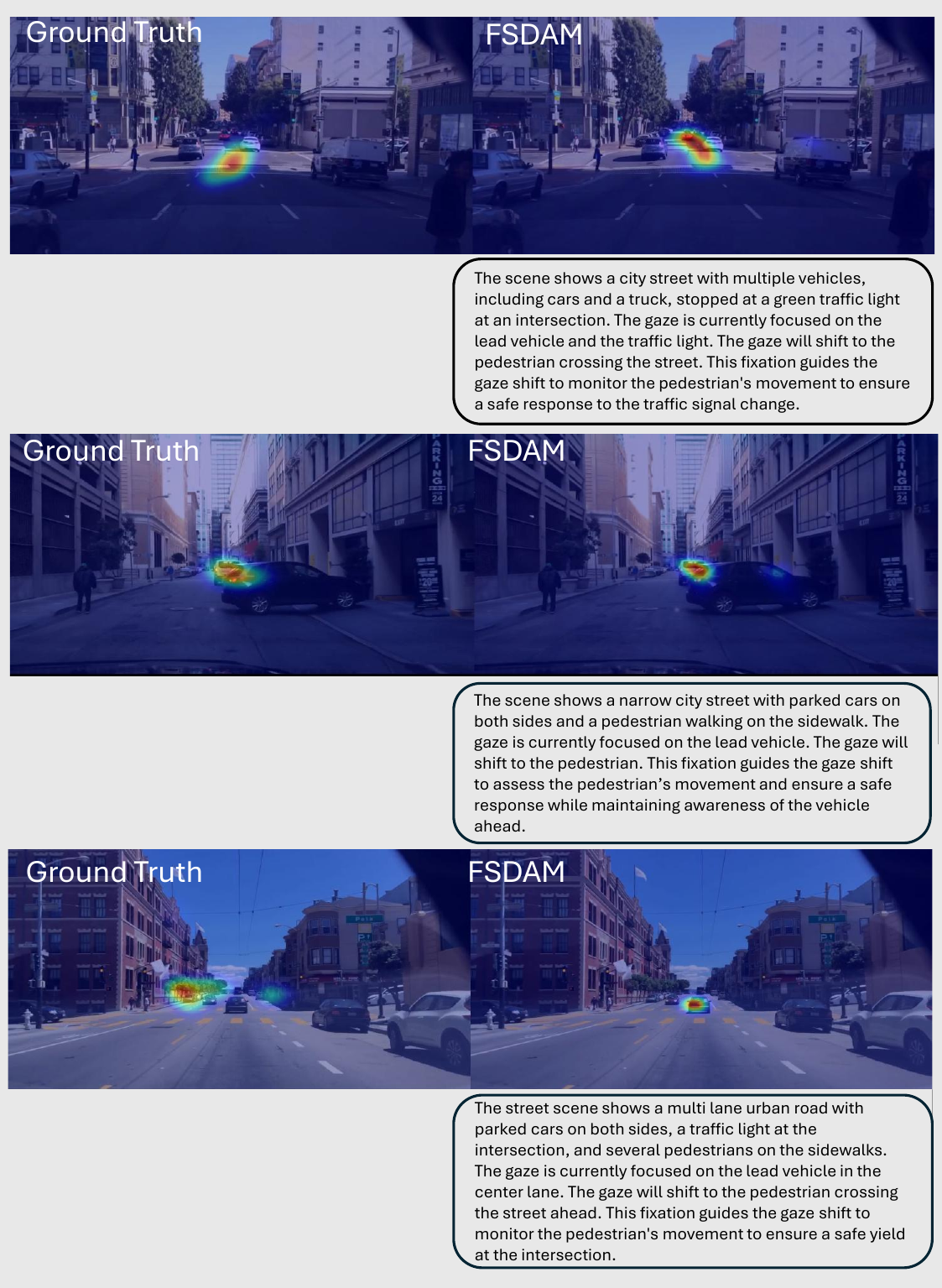}
    \caption{\textbf{FSDAM predictions on BDD-A test set.} Ground truth gaze (left) vs. FSDAM predictions (right) with generated captions. Our model accurately identifies safety-critical regions and generates coherent explanations across diverse urban scenarios: intersections with pedestrians, narrow streets with parked vehicles, and multi-lane roads with complex traffic. }
    \label{fig:bdda_qualitative}    
\end{figure*}

We provide comprehensive qualitative results demonstrating FSDAM's generalization on BDD-A test set, where ground truth captions are unavailable. Figure~\ref{fig:bdda_qualitative} shows representative examples across diverse urban scenarios.

FSDAM's gaze predictions closely match ground truth patterns across all examples, consistently focusing on the most immediate hazards. In the top row, both ground truth and prediction concentrate on the intersection ahead with approaching traffic. The middle example shows accurate attention on the lead vehicle while maintaining awareness of the pedestrian on the sidewalk. The bottom row demonstrates proper focus on the pedestrian crossing area at the intersection. The spatial distribution and intensity of predicted heatmaps align well with ground truth, validating our few-shot learning approach.

The generated captions remain semantically coherent, and even without corresponding ground-truth captions, their validity is reinforced by the qualitative consistency observed between caption content and gaze-attention heatmaps. Each follows our four-sentence structure: scene

\begin{figure*}[t]
    \vspace*{\fill}
    \centering
    \includegraphics[width=0.9\textwidth]{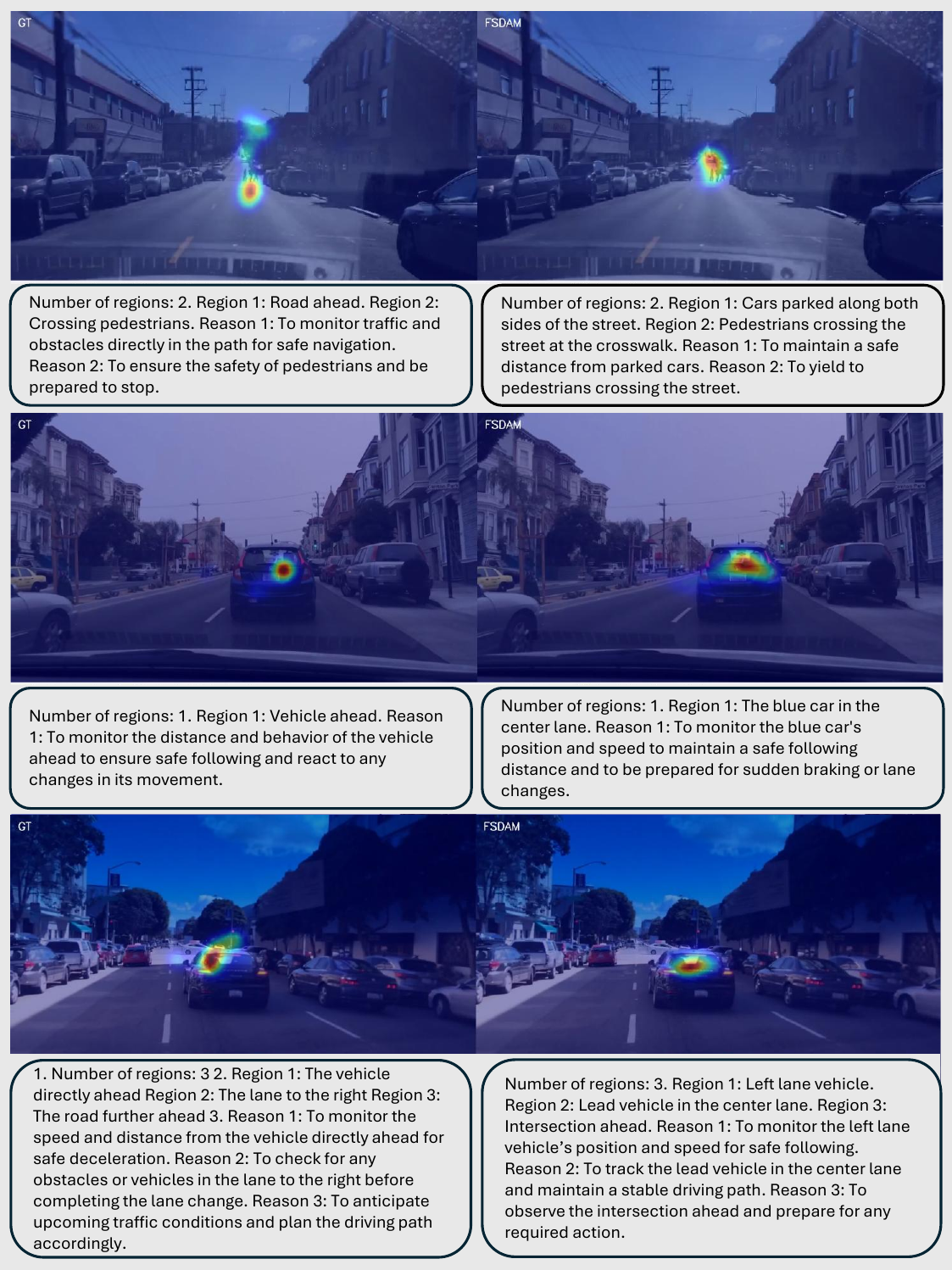}
    \caption{\textbf{Zero-shot FSDAM predictions on W3D\cite{zhou2025where}.} Ground truth annotations (left) and FSDAM predictions (right) highlight the same attention targets and safety cues across diverse scenes. FSDAM generalizes to W3D without any dataset specific training and produces gaze and reasoning patterns consistent with the labeled regions.}
    \label{fig:w3d}
    \vspace*{\fill}
\end{figure*}

\clearpage

description, current fixation, anticipated shift, and safety rationale. Notably, the model correctly identifies gaze targets from its own predictions and provides appropriate driving context. For instance, when detecting pedestrians, it generates ``monitor the pedestrian's movement to ensure a safe yield" rather than generic responses. This consistency across varied scenarios from narrow streets to complex intersections demonstrates that our temporal mining captures fundamental attention patterns that generalize beyond the training distribution.

\subsection{Cross-data generalization on W3D dataset}
We evaluate FSDAM's zero-shot transfer to W3D, which provides ground truth captions alongside gaze annotations. Figure~\ref{fig:w3d} shows representative examples comparing our predictions with W3D's region-based annotations.

W3D uses a distinctive region-numbering format ("Number of regions: 2. Region 1: Road ahead...") that differs fundamentally from FSDAM's output structures. Despite this, both output structures consistently identify identical attention targets across all examples. In pedestrian scenarios, both output emphasize crosswalk regions; in vehicle-following situations, they consistently track the lead vehicle; and at intersections, they distribute attention across several possible conflict zones.

The gaze heatmaps show strong spatial alignment, with FSDAM accurately predicting attention concentration at the same locations as ground truth.

The semantic correspondence extends to safety reasoning despite surface-level differences, where W3D states ``ensure the safety of pedestrians and be prepared to stop", FSDAM generates ``yield to pedestrians crossing the street"—expressing the same defensive driving principle. This alignment emerges without any W3D-specific training, suggesting that our 90-sample curation captures fundamental attention patterns that generalize across annotation protocols and datasets. Strong zero-shot performance validates our temporal mining approach for few-shot learning in safety-critical applications. 

\subsection{Failure Cases}
To better characterize the limitations of FSDAM, we qualitatively analyzed failure cases along two axes: \textit{imperfect visual prioritization} and \textit{cross-modal inconsistency}. The first refers to cases in which the predicted attention map remains partially grounded in the scene but misprioritizes the most behaviorally relevant cue or becomes overly diffuse under visual clutter. The second refers to cases in which the generated explanation is plausible at the language level but is not tightly supported by the attended visual evidence. 
This taxonomy is useful because it distinguishes failures of spatial allocation from cross-modal inconsistency between the gaze prediction and the reasoning, which need not occur simultaneously. In several cases, the model identifies the correct high-level maneuver context while still allocating attention to suboptimal 
regions. In others, the attention map and explanation each appear locally plausible, yet they do not support the same driving rationale when considered jointly. These examples therefore help clarify not only where FSDAM fails, but also how those failures arise.

\begin{figure*}[t]
    \centering
    \includegraphics[width=\textwidth]{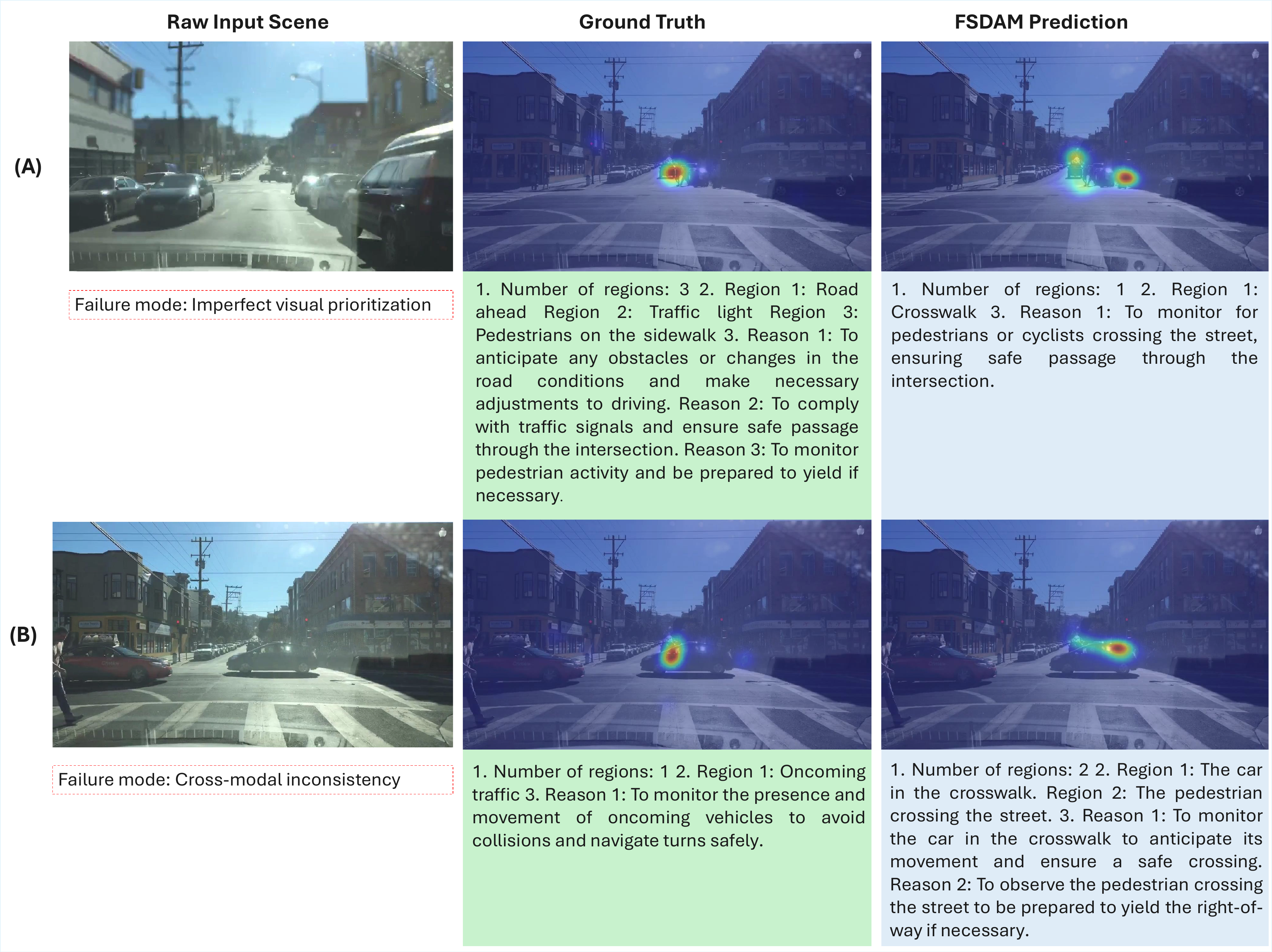}
    \caption{
    Qualitative failure cases of FSDAM.
    \textbf{(A) Imperfect visual prioritization:} the model partially captures the crossing context but also assigns substantial attention to a visually salient right-side object, diluting focus on the primary crossing-related cue.
    \textbf{(B) Cross-modal inconsistency:} the explanation emphasizes crossing-related hazards, whereas the predicted attention is not aligned with the reference oncoming-conflict region.
    }
    \label{fig:failure_1}
\end{figure*}

Figure~\ref{fig:failure_1} shows two representative failures. In Fig.~\ref{fig:failure_1}(A), the model attends to part of the relevant crossing region but also exhibits spurious allocation to a secondary salient object, indicating imperfect prioritization within the scene.This suggests that the model is sensitive to conspicuity even when a more behaviorally important target is present nearby. The error is therefore not a complete loss of scene understanding, but a failure to rank competing cues correctly. Such cases are especially important in 
urban scenes, where multiple salient objects co-occur but only a subset directly constrains the driving decision. In Fig.~\ref{fig:failure_1}(B), the error is cross-model inconsistency: although the generated explanation refers to plausible crossing-related entities, the predicted attention does not support the same hazard configuration as the reference map. This suggests that explanation quality can remain superficially plausible even when visual grounding is incomplete. More broadly, the example indicates that textual plausibility alone is not sufficient evidence of faithful reasoning.

\begin{figure*}[t]
    \centering
    \includegraphics[width=\textwidth]{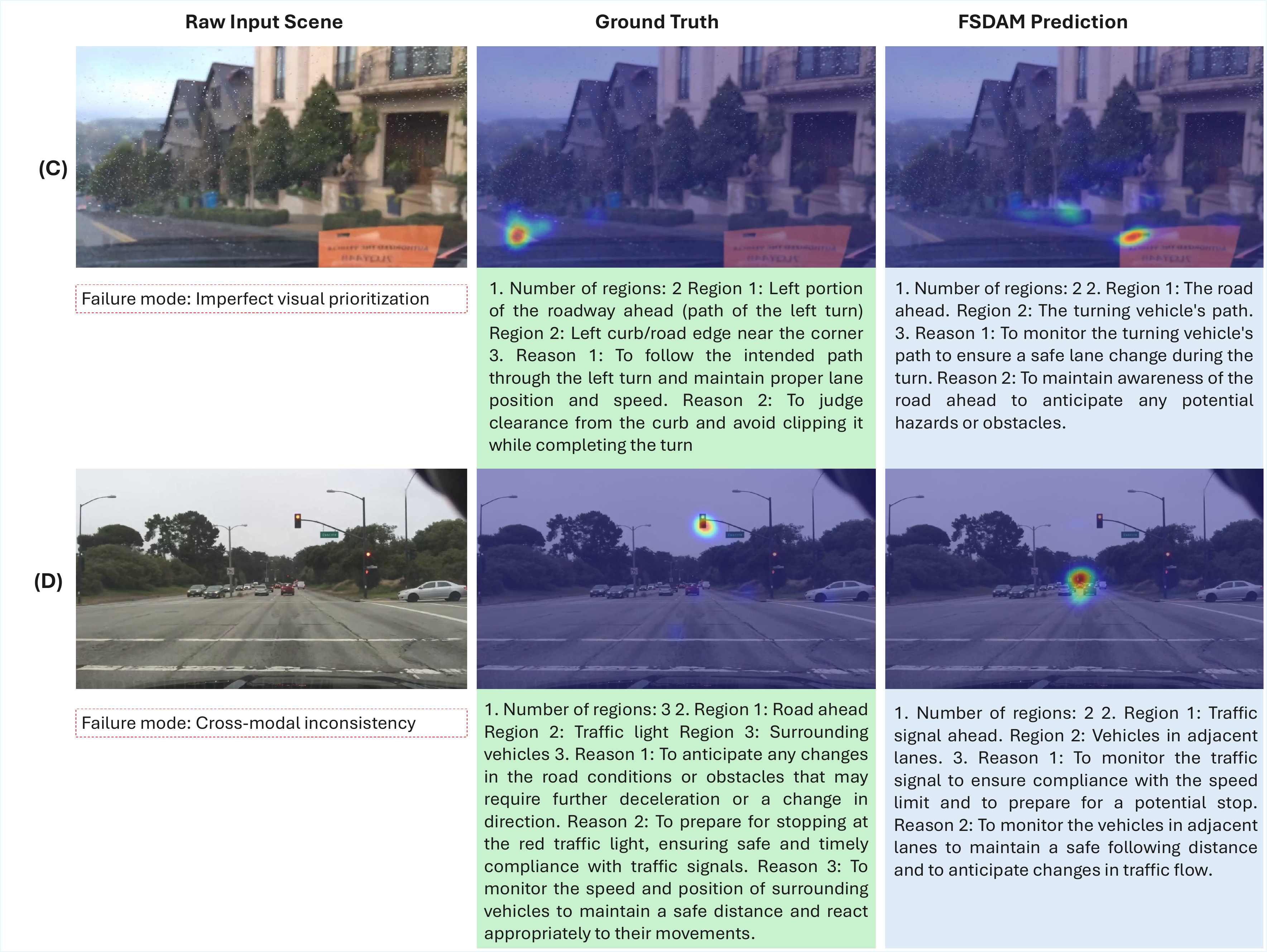}
    \caption{
    Qualitative failure cases of FSDAM.
    \textbf{(C) Imperfect visual prioritization:} during a turning maneuver, attention is diverted toward a foreground dashboard artifact 
    rather than the turn path and curb-clearance region emphasized by the reference map.
    \textbf{(D) Cross-modal inconsistency:} the explanation mentions the traffic signal and surrounding vehicles, but the predicted attention
    is concentrated more strongly on the lead vehicle and forward roadway than on the signal highlighted in the reference.
    }
    \label{fig:failure_2}
\end{figure*}

Figure~\ref{fig:failure_2} demonstrates that these failure modes recur in visually distinct settings. In Fig.~\ref{fig:failure_2}(C), the dominant error is spatial misprioritization caused by foreground saliency. Here, the model is distracted by an image-plane artifact that is visually prominent but irrelevant to the maneuver, revealing limited robustness to nuisance saliency. This type of error is noteworthy because it arises even when the semantically relevant cue is spatially localized and structurally simple. It suggests that improved 
suppression of foreground bias or stronger supervision on maneuver-critical regions may be beneficial. In Fig.~\ref{fig:failure_2}(D), the model again produces an explanation that is semantically plausible but only partially supported by the attended region. The explanation refers to the traffic signal and nearby traffic, whereas the heatmap is dominated by the lead vehicle and forward roadway. Together, these cases indicate that FSDAM remains sensitive both to irrelevant saliency and to incomplete grounding between attention prediction and textual reasoning.

\begin{figure*}[t]
    \centering
    \includegraphics[width=\textwidth]{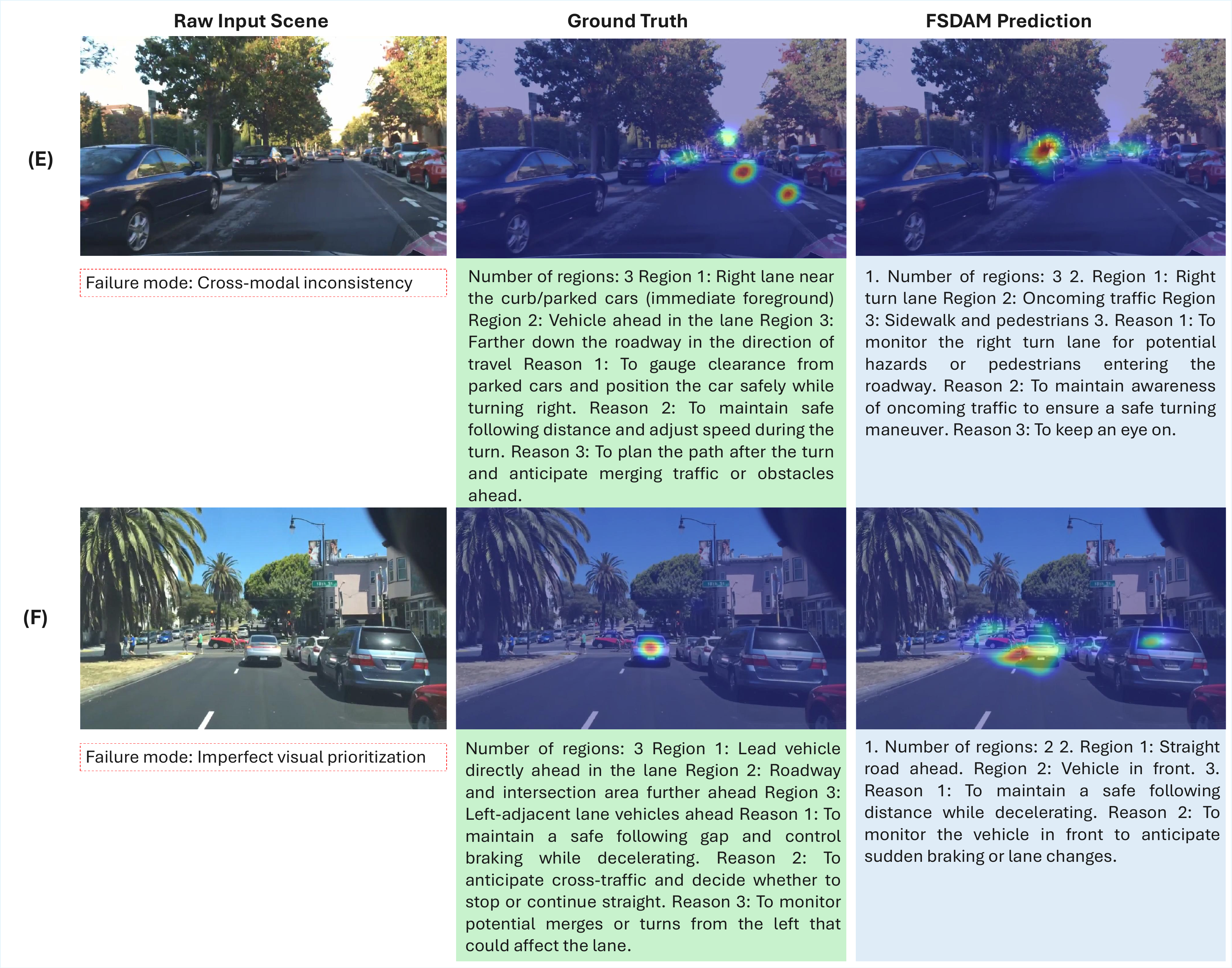}
    \caption{
    Qualitative failure cases of FSDAM.
    \textbf{(E) Cross-modal inconsistency:} in this right-turn scene, the explanation refers to maneuver-relevant hazards, but the attended 
    regions do not align with the reference cues associated with curb clearance, lead-vehicle monitoring, and downstream path planning.
    \textbf{(F) Imperfect visual prioritization:} in a cluttered multi-agent scene, the model captures the general forward driving 
    context but distributes attention too broadly across nearby agents and scene elements relative to the more concentrated reference map.
    }
    \label{fig:failure_3}
\end{figure*}

Figure~\ref{fig:failure_3} highlights two further limitations. In Fig.~\ref{fig:failure_3}(E), the mismatch is again cross-modal: the explanation describes relevant turning hazards, but the visual evidence emphasized by the model does not correspond to the same set of cues. This failure is informative because the explanation is not nonsensical; rather, it is insufficiently grounded in the actual attended regions. In Fig.~\ref{fig:failure_3}(F), the prediction remains broadly grounded in the correct forward scene but loses precision under clutter, 
yielding a diffuse allocation over multiple agents and scene elements. The error here is therefore one of concentration rather than semantic collapse. This distinction matters, since it suggests that part of the remaining performance gap may stem from uncertainty calibration or attention sharpening rather than from missing scene semantics altogether. It also indicates that cluttered multi-agent scenes remain a particularly challenging regime for the model.

Overall, these examples show that FSDAM failures arise primarily from two sources: inaccurate prioritization among competing visual cues and 
incomplete alignment between attention maps and generated explanations. Across cases, the model often captures the global maneuver context 
correctly, but does not always localize the most behaviorally critical cue with sufficient precision. The qualitative evidence also suggests 
that explanation generation can abstract the scene at the correct semantic level while still failing to remain faithful to the underlying 
attended evidence. This gap between semantic plausibility and visual faithfulness is especially important for explainable driving models, 
where interpretability is intended to reflect the model's actual decision basis. From a modeling perspective, the observed errors point 
toward three concrete directions: improved suppression of nuisance saliency, stronger supervision for maneuver-critical cue selection, and 
tighter coupling between explanation generation and attention prediction. Taken together, the failure analysis suggests that the main 
challenge is not recognizing the scene at a coarse level, but resolving which cue should dominate attention and how that cue should be 
verbalized consistently.

\section{Additional Quantitative Analysis}

\subsection{Extension of Table 3: Adding LLada in Few-Shot Setting}

To provide a more rigorous comparison, we extend Table~3 of the main paper by 
including LLada~\cite{zhou2025where} fine-tuned under the same few-shot regime 
as our method. Specifically, LLada$^\dagger$ is fine-tuned on the identical 
90-sample BDD-A subset used to train FSDAM, ensuring a controlled and fair 
comparison. As shown in Table~\ref{tab:w3d_caption_comparison_}, LLada$^\dagger$ 
under the few-shot setting performs substantially below its fully-trained 
counterpart, highlighting the difficulty of adapting language models for driving to extremely limited supervision. In contrast, FSDAM consistently achieves 
competitive performance across all three scenario categories despite using the 
same limited training data, demonstrating the effectiveness of our dual-pathway 
architecture and blur-gap regularization in low-data regimes.

\begin{table*}[!t]
\centering
\caption{Comparison of captioning performance across driving scenarios on W3D dataset. 
Fully-trained baselines use the original W3D training data, 
Zero-shot and ICL models require no fine-tuning, 
and our FSDAM and few-shot baselines are trained on a 90-sample BDD-A subset. 
Higher is better.}
\resizebox{\textwidth}{!}{
\begin{tabular}{l|l|cccc|cccc|cccc}
\toprule
\multirow{2}{*}{\textbf{Method}} &
\multirow{2}{*}{\textbf{Training Regime}} &
\multicolumn{4}{c|}{\textbf{Normal Driving}} &
\multicolumn{4}{c|}{\textbf{Safety-Critical Situation}} &
\multicolumn{4}{c}{\textbf{Traffic Accident}} \\
\cmidrule(lr){3-14}
 &  & BLEU & METEOR & ROUGE & CIDEr-R 
    & BLEU & METEOR & ROUGE & CIDEr-R 
    & BLEU & METEOR & ROUGE & CIDEr-R \\
\midrule
\multicolumn{14}{l}{\textit{Fully Trained on W3D}} \\
\midrule
GazeXplain\textsuperscript{*}~\cite{chen2024gazexplainlearningpredictnatural} & Full-data (W3D) [$\sim$70k samples] & 0.31 & 0.30 & 0.22 & 0.42 & 0.19 & 0.29 & 0.37 & \underline{0.55} & 0.17 & 0.20 & \underline{0.44} & 0.66 \\
LLada\textsuperscript{*}~\cite{zhou2025where} & Full-data (W3D) [$\sim$70k samples] & \textbf{0.44} & \underline{0.36} & \textbf{0.58} & \textbf{0.96} & \textbf{0.44} & \textbf{0.38} & \textbf{0.59} & \textbf{1.23} & \textbf{0.38} & \underline{0.32} & \textbf{0.52} & \textbf{1.00} \\
\midrule
\multicolumn{14}{l}{\textit{Zero-shot and In-Context Models}} \\
\midrule
Qwen-VL~\cite{bai2025qwen25vltechnicalreport} & Zero-shot (no training) 
& 0.10 & 0.19 & 0.28 & 0.34 & 0.19 
& 0.21 & 0.29 & 0.13 & 0.08 
& 0.21 & 0.29 & 0.12 \\

LLaVA~\cite{10.5555/3666122.3667638} & Zero-shot (no training) 
& 0.12 & 0.14 & 0.23 & 0.35 
& 0.13 & 0.19 & 0.11 & 0.10 
& 0.17 & 0.26 & 0.19 & 0.13 \\

Qwen-VL~\cite{bai2025qwen25vltechnicalreport} & In-context learning (no fine-tuning) 
& 0.13 & 0.18 & 0.22 & 0.36 & 0.21 
& 0.17 & 0.30 & 0.23 & 0.12 
& 0.24 & 0.33 & 0.15 \\

\midrule
\multicolumn{14}{l}{\textit{Few-shot Learning (BDD-A 90 samples)}} \\
\midrule
DeepGazeI\cite{kümmerer2015deepgazeiboosting} + LLaVA & Few-shot & 0.12 & 0.23 & 0.28 & 0.14 & 0.13 & 0.22 & 0.30 & 0.18 & 0.15 & 0.21 & 0.31 & 0.17 \\
DeepGazeIIE\cite{linardos2021deepgazeiiecalibratedprediction} + LLaVA & Few-shot & 0.11 & 0.18 & 0.26 & 0.17 & 0.11 & 0.20 & 0.32 & 0.13 & 0.11 & 0.19 & 0.34 & 0.10 \\
MLNet\cite{8356626} + LLaVA & Few-shot & 0.13 & 0.19 & 0.27 & 0.31 & 0.26 & 0.20 & 0.32 & 0.12 & 0.13 & 0.18 & 0.33 & 0.28 \\
LLada$^\dagger$~\cite{zhou2025where} & Few-shot & 0.15 & 0.18 & 0.19 & 0.12 & 0.11 & 0.16 & 0.16 & 0.10 & 0.13 & 0.18 & 0.19 & 0.17 \\
\textbf{FSDAM (Ours)} & Few-shot & \underline{0.42} & \textbf{0.37} & \underline{0.48} & \underline{0.83} & \underline{0.35} & \underline{0.33} & \underline{0.46} & 0.47 & \underline{0.33} & \textbf{0.35} & 0.34 & \underline{0.84} \\
\bottomrule
\end{tabular}}
\label{tab:w3d_caption_comparison_}
\end{table*}

\begin{table}[t]
\centering
\small
\caption{Variance analysis of FSDAM on BDD-A across three random seeds (90 training samples). Results are reported as mean $\pm$ standard deviation.}
\label{tab:variance}
\resizebox{\columnwidth}{!}{
\begin{tabular}{lcccccc}
\toprule
Method & CC$\uparrow$ & KL$\downarrow$ & SIM$\uparrow$ & AUC-J$\uparrow$ & AUC-B$\uparrow$ & NSS$\uparrow$ \\
\midrule
FSDAM (Seed 1) & 0.60 & 1.13 & 0.43 & 0.96 & 0.91 & 4.10 \\
FSDAM (Seed 2) & 0.58 & 1.17 & 0.42 & 0.95 & 0.90 & 3.98 \\
FSDAM (Seed 3) & 0.61 & 1.11 & 0.44 & 0.96 & 0.91 & 4.15 \\
\midrule
\textbf{Mean $\pm$ Std} & \textbf{0.60 $\pm$ 0.01} & \textbf{1.14 $\pm$ 0.03} & \textbf{0.43 $\pm$ 0.01} & \textbf{0.96 $\pm$ 0.01} & \textbf{0.91 $\pm$ 0.01} & \textbf{4.08 $\pm$ 0.09} \\
\bottomrule
\end{tabular}}
\end{table}

\subsection{Variance Analysis Under Random Support Sampling}
To evaluate the stability of FSDAM under different support-set compositions, we report performance on BDD-A across three random seeds and 
report mean $\pm$ standard deviation for each metric. Table~\ref{tab:variance} shows that performance remains consistent across seeds, 
with only small fluctuations across all metrics. This indicates that FSDAM is not sensitive to the specific random selection of few-shot 
support samples, and that the observed improvements do not depend on a particularly favorable support-set draw. These results confirm 
that FSDAM's performance is reproducible and robust to support-set composition in the low-data regime.

\subsection{Expanded Few-Shot and Robustness Analysis}

To further analyze the data efficiency and robustness of FSDAM, we conduct additional quantitative experiments beyond the main paper. These experiments examine how performance scales with the number of training samples, whether the curated few-shot set depends heavily on any single driving category, how each structured caption component contributes to gaze prediction, and whether FSDAM benefits from additional target-domain supervision.

\paragraph{Expanded few-shot scaling.}
Table~\ref{tab:expanded_fewshot} reports an expanded few-shot scaling study on BDD-A using $N \in \{8,16,24,32,40,50,70,90\}$ training samples. Performance improves rapidly as the number of annotated samples increases. With only 8 samples, FSDAM achieves KL/CC/SIM/NSS of $1.74/0.41/0.27/2.32$. With 40 samples, performance improves to $1.17/0.58/0.43/4.09$, already approaching the 90-sample result of $1.13/0.60/0.43/4.10$. This indicates that most gains are obtained by approximately 40 samples, while additional examples mainly refine correlation and peak localization.

\begin{table}[t]
\centering
\caption{Expanded few-shot scaling on BDD-A. Performance improves rapidly up to 40 samples and then saturates.}
\label{tab:expanded_fewshot}
\begin{tabular}{c|cccccc}
\toprule
$N$ & KL$\downarrow$ & CC$\uparrow$ & SIM$\uparrow$ & NSS$\uparrow$ & AUC-B$\uparrow$ & AUC-J$\uparrow$ \\
\midrule
8  & 1.74 & 0.41 & 0.27 & 2.32 & 0.890 & 0.910 \\
16 & 1.73 & 0.42 & 0.28 & 2.37 & 0.890 & 0.910 \\
24 & 1.39 & 0.53 & 0.37 & 3.64 & 0.890 & 0.960 \\
32 & 1.38 & 0.54 & 0.35 & 3.52 & 0.913 & 0.962 \\
40 & 1.17 & 0.58 & 0.43 & 4.09 & 0.907 & 0.964 \\
50 & 1.16 & 0.58 & 0.41 & 4.07 & 0.917 & 0.955 \\
70 & 1.15 & 0.58 & 0.41 & 4.09 & 0.921 & 0.960 \\
90 & 1.13 & 0.60 & 0.43 & 4.10 & 0.917 & 0.960 \\
\bottomrule
\end{tabular}
\end{table}

\paragraph{Category dropout robustness.}
We next evaluate whether FSDAM depends disproportionately on any single scenario category in the 90-sample training set. Table~\ref{tab:category_dropout} reports results after removing one category at a time. Performance remains stable under most category removals. For example, removing normal-driving samples changes KL/NSS only from $1.13/4.10$ to $1.15/3.96$. The largest degradation occurs when unsignalized or construction-related samples are removed, suggesting that these categories contain diverse attention-transition cues useful for generalization. Overall, the results show that FSDAM is not driven by one dominant category and that the curated few-shot set provides robust supervision across driving scenarios.

\begin{table*}[t]
\centering
\caption{Category dropout analysis. Each column removes one scenario category from the 90-sample training set.}
\label{tab:category_dropout}
\small
\setlength{\tabcolsep}{4pt}
\begin{tabular}{l|c|c|c|c|c|c|c|c}
\toprule
Metric
& None
& \shortstack{Normal\\Driving}
& \shortstack{Construction\\Zone}
& \shortstack{Imminent\\Hazard}
& \shortstack{Lane\\Change}
& \shortstack{Pedestrian\\Crossing}
& \shortstack{Signalized\\Intersection}
& \shortstack{Unsignalized\\Intersection} \\
\midrule
KL$\downarrow$  & 1.13 & 1.15 & 1.21 & 1.13 & 1.18 & 1.13 & 1.16 & 1.24 \\
CC$\uparrow$    & 0.60 & 0.59 & 0.58 & 0.60 & 0.58 & 0.60 & 0.59 & 0.57 \\
SIM$\uparrow$   & 0.43 & 0.42 & 0.40 & 0.45 & 0.42 & 0.44 & 0.43 & 0.39 \\
NSS$\uparrow$   & 4.10 & 3.96 & 3.85 & 4.16 & 3.91 & 4.09 & 4.06 & 3.77 \\
\bottomrule
\end{tabular}
\end{table*}
\paragraph{Structured caption component ablation.}
Table~\ref{tab:caption_structure_ablation} studies the effect of each structured reasoning component. Using only scene context and current gaze descriptions gives KL/CC/NSS of $1.27/0.53/3.71$. Adding the next-focus component improves KL and CC to $1.24$ and $0.54$, respectively. The full structured representation, which additionally includes the causal rationale, further improves KL/CC/NSS to $1.13/0.60/4.10$. These results indicate that short-horizon transition supervision and causal reasoning provide useful semantic regularization for gaze prediction, even though future frames are not used at inference.

\begin{table}[t]
\centering
\caption{Ablation of structured caption components. $C_s$, $C_c$, $C_n$, and Full denote scene context, current gaze, next focus, and the complete structure including causal rationale.}
\label{tab:caption_structure_ablation}
\begin{tabular}{l|ccc}
\toprule
Metric & $C_s+C_c$ & $+C_n$ & Full \\
\midrule
KL$\downarrow$  & 1.27 & 1.24 & 1.13 \\
CC$\uparrow$   & 0.53 & 0.54 & 0.60 \\
SIM$\uparrow$  & 0.43 & 0.43 & 0.43 \\
NSS$\uparrow$  & 3.71 & 3.73 & 4.10 \\
\bottomrule
\end{tabular}
\end{table}

\paragraph{Additional W3D training-scale experiment.}
Although FSDAM is designed for few-shot transfer, we also test whether the architecture benefits from additional target-domain supervision. Table~\ref{tab:w3d_training_scale} compares captioning performance when training on 100 W3D samples versus 13K W3D samples. Increasing target-domain supervision improves BLEU from $0.28$ to $0.37$, METEOR from $0.31$ to $0.38$, ROUGE from $0.25$ to $0.32$, and CIDEr from $0.29$ to $0.49$. This suggests that FSDAM retains strong low-data efficiency while still benefiting from larger target-domain training sets.

\begin{table}[t]
\centering
\caption{Additional W3D training-scale experiment. FSDAM benefits from larger target-domain supervision.}
\label{tab:w3d_training_scale}
\begin{tabular}{l|cc}
\toprule
Metric & 100 samples & 13K samples \\
\midrule
BLEU    & 0.28 & 0.37 \\
METEOR  & 0.31 & 0.38 \\
ROUGE   & 0.25 & 0.32 \\
CIDEr   & 0.29 & 0.49 \\
\bottomrule
\end{tabular}
\end{table}

\paragraph{Human evaluation of generated explanations.}
Automatic captioning metrics do not fully capture whether generated explanations are causally meaningful, visually grounded, and relevant to driving. We therefore conduct a human evaluation on 30 generated explanations using three experienced driver raters. Scores are reported on a 1--5 scale. As shown in Table~\ref{tab:human_eval}, the generated ``Why'' explanations obtain $3.90 \pm 0.44$ for causal correctness, $4.53 \pm 0.50$ for visual grounding, and $4.66 \pm 0.48$ for driving relevance. These results complement the automatic captioning metrics and indicate that FSDAM produces explanations that are generally grounded in visible scene evidence and relevant to the driving context.

\begin{table}[t]
\centering
\caption{Human evaluation of generated ``Why'' explanations on 30 samples by three experienced driver raters. Scores use a 1--5 scale.}
\label{tab:human_eval}
\begin{tabular}{l|c}
\toprule
Criterion & Score \\
\midrule
Causal correctness & $3.90 \pm 0.44$ \\
Visual grounding & $4.53 \pm 0.50$ \\
Driving relevance & $4.66 \pm 0.48$ \\
\bottomrule
\end{tabular}
\end{table}

\paragraph{Per-category gaze analysis.}
Finally, Table~\ref{tab:per_category_gaze} reports gaze prediction performance across scenario types. FSDAM performs similarly on normal driving and safety-critical situations, with CC/SIM of $0.58/0.47$ and $0.59/0.48$, respectively. Performance is lower on traffic accident scenes, where CC/SIM drops to $0.44/0.36$. This is expected because accident scenes often contain higher visual clutter, multiple plausible attention targets, and greater inter-driver gaze variability.

\begin{table}[t]
\centering
\caption{Per-category gaze prediction breakdown. Accident scenes are more challenging due to higher scene complexity and gaze variability.}
\label{tab:per_category_gaze}
\begin{tabular}{l|cc}
\toprule
Category & CC$\uparrow$ & SIM$\uparrow$ \\
\midrule
Normal driving & 0.58 & 0.47 \\
Safety-critical situation & 0.59 & 0.48 \\
Traffic accident & 0.44 & 0.36 \\
\bottomrule
\end{tabular}
\end{table}
\subsection{Key-Frame Selection Ablation}
To separate the effect of data curation from architecture, we vary only how the 90 training
frames are chosen, keeping the model, hyperparameters, and eight-category coverage fixed. We 
compare random BDD-A frames, high-saliency frames, and our KL-mined transition frames. As 
the captions are generated for KL-mined pairs only, this ablation is gaze-only. 
Table~\ref{tab:selection_ablation} shows that KL-mined selection gives the best CC and SIM, indicating that transition-aware curation improves few-shot gaze prediction.
\begin{table}[t]
\centering
\small
\caption{Key-frame selection ablation on BDD-A (90 samples, gaze-only, mean $\pm$ std over 3 seeds).}
\label{tab:selection_ablation}
\begin{tabular}{lcc}
\toprule
Selection Strategy & CC$\uparrow$ & SIM$\uparrow$ \\
\midrule
Random frames & 0.54 $\pm$ 0.03 & 0.39 $\pm$ 0.02 \\
High-saliency frames & 0.56 $\pm$ 0.02 & 0.40 $\pm$ 0.02 \\
\midrule
\textbf{KL-mined (Ours)} & \textbf{0.60 $\pm$ 0.01} & \textbf{0.43 $\pm$ 0.01} \\
\bottomrule
\end{tabular}
\end{table}

%
%
\clearpage

\end{document}